%% file: iclr2027_conference.tex
\documentclass{article} 
\usepackage{iclr2027_conference,times}

\input{math_commands.tex}

\usepackage{hyperref}
\usepackage{url}
\usepackage{tikz}
\usetikzlibrary{arrows.meta,positioning,calc,backgrounds,shapes.geometric,decorations.pathreplacing,fit}
\usepackage{pgfplots}
\usepgfplotslibrary{groupplots}
\pgfplotsset{compat=1.18}
\usepackage{wrapfig}
\usepackage{algorithm}
\usepackage{algorithmic}
\usepackage{float}
\usepackage{amssymb}
\usepackage{amsthm}
\usepackage{placeins}
\usepackage{flafter}
\usepackage{graphicx}
\usepackage{booktabs}

\newtheorem{proposition}{Proposition}

\title{Most-Recent Anchoring with Recurrent Ordering for Time Series Forecasting}

\author{Jung Min Choi$^{*}$, Ngoc Son Le$^{*}$, Ibram Abdelmalak, \\
\textbf{Vijaya Krishna Yalavarthi \& Lars Schmidt-Thieme} \\
ISMLL, University of Hildesheim \\
VWFS Data Analytics Research Center (VWFS-DARC), Hildesheim, Germany \\
\texttt{choi@ismll.de}, \texttt{sle@ismll.de}, \texttt{abdelmalak@ismll.de}, \\
\texttt{yalavarthi@ismll.de}, \texttt{schmidt-thieme@ismll.de}
}

\iclrfinalcopy
\begin{document}
\raggedbottom

\maketitle
{\renewcommand{\thefootnote}{\fnsymbol{footnote}}\footnotetext[1]{Equal contribution.}}

\begin{abstract}
\input{sections/abstract}
\end{abstract}

\FloatBarrier
\input{sections/introduction}
\input{sections/related_work}
\input{sections/method}
\input{sections/experiments}
\input{sections/conclusion}

\subsection*{AI use statement}

Generative AI tools were used to draft parts of the manuscript, revise the
writing, and help find related work. The authors checked all suggested
literature against the original sources and manually reviewed all AI-assisted
text. Generative AI tools were not used to generate datasets, develop the
model or its theoretical claims, design the experiments, implement the method,
preprocess the data, or interpret the results. The authors take responsibility
for the final paper, including its text, claims, citations, and artifacts.

\subsection*{Reproducibility statement}

The MARO architecture and its implementation-level computation are specified
in Section~\ref{sec:method}, including patching, channel mixing, calendar
conditioning, recurrent refinement, checkpoint fusion, and forecast decoding.
Section~\ref{sec:experiments} describes the training objective and experimental
protocols. Appendix~\ref{app:data} gives the benchmark statistics, chronological
splits, training-only standardization, sliding-window construction, and RevIN
preprocessing, while Appendix~\ref{app:metrics} defines the evaluation measures.
The hyperparameter-search budget and complete search space are reported in
Appendix~\ref{app:space}, and the selected long-term and PEMS configurations are
reported in Appendix~\ref{app:selected}. The measurement protocol and hardware
for the efficiency experiments are given in Appendix~\ref{app:efficiency}.
The anonymous GitHub repository supplied with the submission contains the
training and evaluation code and the experimental configurations needed to
reproduce the reported results.

\bibliography{iclr2027_conference}
\bibliographystyle{iclr2027_conference}

\appendix
\input{sections/appendix}

\end{document}

%% file: math_commands.tex
\usepackage{amsmath,amsfonts,bm}

\def\eqref#1{equation~\ref{#1}}

\def\1{\bm{1}}

\DeclareMathAlphabet{\mathsfit}{\encodingdefault}{\sfdefault}{m}{sl}
\SetMathAlphabet{\mathsfit}{bold}{\encodingdefault}{\sfdefault}{bx}{n}



%% file: sections/abstract.tex
Long-term forecasting models commonly process all patches in a look-back window
using the same fixed stack. Older contextual patches and recent evidence
therefore receive the same computational depth. Yet the information closest
to the forecast and the more distant context do not contribute equally.
Uniform processing leaves this distinction unexpressed in the architecture.
We propose MARO, a \textbf{M}ost-Recent \textbf{A}nchoring
with \textbf{R}ecurrent \textbf{O}rdering model that processes the look-back window
from the most recent patch to the oldest. The most recent patch serves as the
anchor. It initializes
the latent state and conditions each subsequent step, so older patches are folded
into a representation that remains centered on recent evidence. A single shared
module is reused at every step, so extending the scan further into the past
introduces no additional parameters. Intermediate states retained during the
scan allow the forecast head to weigh short and long portions of the history
separately. This expresses recency through the order of recurrent refinement.
Extensive experiments across multiple real-world time series
datasets show that MARO achieves state-of-the-art performance on both long-term
and short-term forecasting tasks. Ablation studies examine the contribution of
the main architectural components.

%% file: sections/introduction.tex
\section{Introduction}
\label{sec:introduction}

\input{sections/fig_schematic}

Time series forecasting (TSF) supports decision-making in domains such as
energy and climate as well as transportation \citep{weather,electricity,pems}.
Long-term TSF has been studied with attention-based \citep{itransformer} and
convolutional \citep{scinet} architectures as well as recurrent
\citep{lin2023segrnn} and MLP-based architectures
\citep{tsmixer,timemixer}. Many non-recurrent models map the
look-back window through a predetermined stack of processing blocks.
Patch-based models first divide the window into tokens and then apply the same
fixed-depth processing to every patch \citep{patchtst,tsmixer2}. Each patch
undergoes the same number of transformations regardless of its distance from
the forecast origin.

Forecasting uses the positions in a look-back window asymmetrically.
Patches adjacent to the forecast boundary often provide the most immediate
evidence. Earlier patches more often supply longer-range context. This
observation motivates anchoring the forecast in the most recent patch and
incorporating progressively older patches as contextual corrections.
Positional information can encode recency, but it does not give patches
different numbers of recurrent refinement steps. A fixed stack still subjects
every patch to the same number of transformations. We use computational depth
to mean the number of recurrent refinement steps through which the information
from a patch is carried.

To test whether reversing the scan provides a useful recency bias across update
mechanisms, we compared recent-to-old and old-to-recent ordering with causal
attention and RNN updates. Recent-to-old begins at the forecast boundary and
moves backward, whereas old-to-recent begins with the earliest patch and moves
forward. Under causal attention, this reversal places recent context in the
visible prefix. Under the RNN update, it initializes the hidden state from the
most recent patch before adding older history. Recent-to-old scanning produced
lower test MSE in all four comparisons in Figure~\ref{fig:scan_direction}, with
reductions of 0.2 and 1.2 percent on ETTh2 and 4.7 and 0.8 percent on Weather.
The consistency across the two mechanisms provides empirical evidence that the
benefit of reversing the scan to recent-to-old is not specific to one update
rule.

\input{sections/fig_scan_direction}

Recurrent-depth computation allows this effective depth to vary.
Recent reasoning systems allocate additional computation at the token or latent
level \citep{cot,coconut,recurrentdepth}. Some iteratively reuse a shared block
without increasing the parameter count \citep{recurrentdepth}.
Applied to a look-back window, the computation assigned to a patch can depend
on its temporal position. Fixed-depth patch mixers do not express this depth
asymmetry.

We propose MARO, a \textbf{M}ost-Recent \textbf{A}nchoring with
\textbf{R}ecurrent \textbf{O}rdering model that scans overlapping patches
from the most recent patch to the oldest
(Figure~\ref{fig:schematic}). The most recent patch initializes the latent state and
serves as a fixed anchor while a shared module progressively integrates earlier
patches as context. MARO retains intermediate states along this scan and combines
them for forecasting, allowing the prediction to draw separately on shorter- and
longer-range portions of the history. The same module is reused at every step,
so incorporating older patches does not add parameters.

Our contributions are threefold.
\begin{itemize}
    \item We identify uniform computational depth as a structural limitation of
    patch-based forecasting and formulate the history as a sequence of recurrent
    refinements organized by recency.
    \item We introduce MARO as a parameter-shared architecture that anchors the
    forecast at the most recent patch and progressively integrates older context.
    This design makes recency explicit in the computation while keeping the
    recurrent parameters independent of the number of patches.
    \item We evaluate MARO across long-term and short-term forecasting benchmarks.
    The results and controlled ablations show when recency-aware computation helps
    and how each architectural component contributes to forecasting accuracy.
    The code is available at \url{https://anonymous.4open.science/r/MARO-7B20/README.md}.
\end{itemize}

%% file: sections/fig_schematic.tex
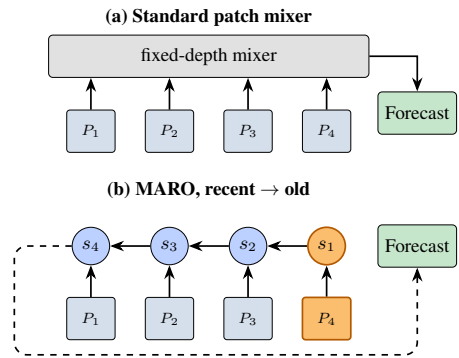
\begin{wrapfigure}{r}{0.42\textwidth}
\centering
\vspace{-8pt}
\definecolor{patchcolor}{RGB}{213,223,236}
\definecolor{anchorcolor}{RGB}{250,190,110}
\definecolor{fusioncolor}{RGB}{198,230,203}
\definecolor{mixercolor}{RGB}{225,225,225}
\definecolor{refinecolor}{RGB}{188,209,255}

\resizebox{\linewidth}{!}{%
\begin{tikzpicture}[>=Stealth,
    patch/.style={rectangle, draw, rounded corners=1.5pt, fill=patchcolor,
        minimum width=0.8cm, minimum height=0.7cm, font=\scriptsize},
    anchorpatch/.style={patch, fill=anchorcolor, draw=orange!70!black, thick},
    outbox/.style={rectangle, draw, rounded corners=2pt, fill=fusioncolor,
        minimum height=0.7cm, minimum width=1.1cm, font=\small, align=center},
    bigbox/.style={rectangle, draw, rounded corners=2pt, fill=mixercolor,
        minimum height=0.7cm, font=\small, align=center},
    state/.style={circle, draw, fill=refinecolor, minimum size=0.62cm,
        font=\small, inner sep=0pt},
    anchorstate/.style={state, fill=anchorcolor, draw=orange!70!black, thick},
]

\def\xa{0}\def\xb{1.3}\def\xc{2.6}\def\xd{3.9}

\node[font=\bfseries\small] at (1.95, 1.88) {(a) Standard patch mixer};

\node[bigbox, minimum width=5.3cm] (mixer) at (1.95, 1.24) {fixed-depth mixer};

\node[patch] (a1) at (\xa,0) {$P_1$};
\node[patch] (a2) at (\xb,0) {$P_2$};
\node[patch] (a3) at (\xc,0) {$P_3$};
\node[patch] (a4) at (\xd,0) {$P_4$};

\foreach \p in {a1,a2,a3,a4}{
    \draw[->, thick] (\p.north) -- (\p.north |- mixer.south);
}

\node[outbox] (fca) at (\xd+1.5, 0.30) {Forecast};
\draw[->, thick] (mixer.east) -| (fca.north);

\node[font=\bfseries\small] at (1.95, -0.98) {(b) MARO, recent $\to$ old};

\node[anchorstate] (s1) at (\xd,-1.9) {$s_1$};
\node[state] (s2) at (\xc,-1.9) {$s_2$};
\node[state] (s3) at (\xb,-1.9) {$s_3$};
\node[state] (s4) at (\xa,-1.9) {$s_4$};

\draw[->, thick] (s1) -- (s2);
\draw[->, thick] (s2) -- (s3);
\draw[->, thick] (s3) -- (s4);

\node[patch] (b1) at (\xa,-3.1) {$P_1$};
\node[patch] (b2) at (\xb,-3.1) {$P_2$};
\node[patch] (b3) at (\xc,-3.1) {$P_3$};
\node[anchorpatch] (b4) at (\xd,-3.1) {$P_4$};

\draw[->, thick] (b4.north) -- (s1.south);
\draw[->, thick] (b3.north) -- (s2.south);
\draw[->, thick] (b2.north) -- (s3.south);
\draw[->, thick] (b1.north) -- (s4.south);

\node[outbox] (fcb) at (\xd+1.5, -1.9) {Forecast};
\draw[->, thick, dashed, rounded corners=6pt]
    (s4.west) -- ++(-0.95,0) -- ++(0,-1.8) -| (fcb.south);

\end{tikzpicture}%
}
\vspace{-4pt}
\caption{Conventional patch mixers transform every patch at the same fixed
depth in panel (a). MARO starts from the most recent patch and recurrently
incorporates older patches in panel (b).}
\label{fig:schematic}
\vspace{-6pt}
\end{wrapfigure}

%% file: sections/fig_scan_direction.tex
\begin{wrapfigure}{r}{0.48\textwidth}
\centering
\vspace{-8pt}
\definecolor{directiongreen}{RGB}{54,150,28}
\definecolor{directiongray}{RGB}{137,136,132}
\definecolor{directionrule}{RGB}{222,222,222}
\definecolor{directionborder}{RGB}{190,190,190}
\setlength{\fboxsep}{4pt}
\fcolorbox{directionborder}{white}{\resizebox{\dimexpr\linewidth-2\fboxsep-2\fboxrule\relax}{!}{%
\begin{tikzpicture}[
    oldpoint/.style={circle, fill=directiongray, draw=none,
        minimum size=4.8pt, inner sep=0pt},
    recentpoint/.style={circle, fill=directiongreen, draw=none,
        minimum size=5.8pt, inner sep=0pt},
    scanarrow/.style={directiongreen, line width=1.0pt,
        -{Stealth[length=1.7mm,width=1.25mm]}},
    rowlabel/.style={font=\small, anchor=west},
    value/.style={font=\scriptsize, text=black!70},
    gain/.style={font=\scriptsize, text=directiongreen!78!black},
    dataset/.style={font=\small\bfseries, text=black!70, anchor=west},
]

\node[dataset] at (0.0,4.00) {ETTh2};
\draw[directionrule, line width=0.45pt] (0.0,3.76) -- (7.25,3.76);

\node[rowlabel] at (0.0,3.10) {Causal attention};
\draw[scanarrow] (3.80,3.10) -- (3.40,3.10);
\node[oldpoint] at (3.80,3.10) {};
\node[recentpoint] at (3.32,3.10) {};
\node[value, anchor=south] at (3.80,3.28) {0.2608};
\node[gain, anchor=north] at (3.32,2.92) {0.2602 ($-0.2\%$)};

\node[rowlabel] at (0.0,2.10) {RNN};
\draw[scanarrow] (6.85,2.10) -- (4.08,2.10);
\node[oldpoint] at (6.85,2.10) {};
\node[recentpoint] at (4.00,2.10) {};
\node[value, anchor=south] at (6.85,2.28) {0.2642};
\node[gain, anchor=north] at (4.00,1.92) {0.2611 ($-1.2\%$)};

\node[dataset] at (0.0,1.26) {Weather};
\draw[directionrule, line width=0.45pt] (0.0,1.02) -- (7.25,1.02);

\node[rowlabel] at (0.0,0.36) {Causal attention};
\draw[scanarrow] (6.95,0.36) -- (3.33,0.36);
\node[oldpoint] at (6.95,0.36) {};
\node[recentpoint] at (3.25,0.36) {};
\node[value, anchor=south] at (6.95,0.54) {0.1551};
\node[gain, anchor=north] at (3.25,0.18) {0.1478 ($-4.7\%$)};

\node[rowlabel] at (0.0,-0.64) {RNN};
\draw[scanarrow] (4.95,-0.64) -- (4.32,-0.64);
\node[oldpoint] at (4.95,-0.64) {};
\node[recentpoint] at (4.24,-0.64) {};
\node[value, anchor=south] at (4.95,-0.46) {0.1512};
\node[gain, anchor=north] at (4.24,-0.82) {0.1500 ($-0.8\%$)};

\node[oldpoint] at (0.70,-1.48) {};
\node[font=\scriptsize, anchor=west, text=black!70] at (0.92,-1.48)
    {Old-to-recent (baseline)};
\node[recentpoint] at (4.12,-1.48) {};
\node[font=\scriptsize, anchor=west, text=directiongreen!78!black]
    at (4.36,-1.48) {Recent-to-old};

\end{tikzpicture}%
}}%
\vspace{-4pt}
\caption{Test MSE under old-to-recent and recent-to-old scanning with causal
attention and RNN updates. Recent-to-old scanning gives lower MSE in all four
comparisons.}
\label{fig:scan_direction}
\vspace{-6pt}
\end{wrapfigure}
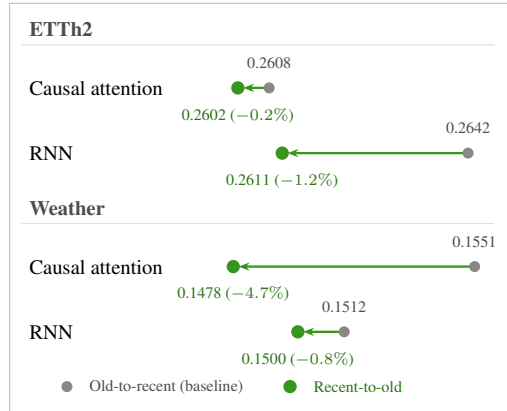

%% file: sections/related_work.tex
\section{Related Work}
\label{sec:related}

\paragraph{Time Series Forecasting.}
Recent forecasting baselines span linear mappings and convolutions as well as
attention and multiscale mixing. DLinear and RLinear use direct linear mappings
\citep{dlinear,rlinear}. SCINet models temporal dynamics through convolution and
interaction \citep{scinet}. iTransformer represents variables as tokens
\citep{itransformer}. PatchTST and Crossformer organize the input into temporal
patches \citep{patchtst,crossformer}. TimeMixer and TimeKAN learn multiscale or
frequency-decomposed representations \citep{timemixer,timekan}. Amplifier
emphasizes low-energy components \citep{amplifier}. CycleNet models periodic
structure \citep{cyclenet}. SRSNet selects among alternative patch
representations \citep{srsnet}. These models form the principal comparison set
in our long-term and short-term experiments. Their architectures differ, but
they process the input through stacks of predetermined depth.

\paragraph{Patching.}
PatchTST divides time series into overlapping temporal segments
\citep{patchtst}. Crossformer additionally captures dependencies between
variables \citep{crossformer}. Subsequent approaches introduce adaptive patch
lengths \citep{hdmixer}. Other methods choose among alternative patch representations
\citep{srsnet} or integrate patching with wavelet analysis \citep{wpmixer}. Some
build and prune graphs among patches \citep{timefilter}. Each patch in these
approaches is generally embedded once before all patch
representations are processed together. MARO instead right-aligns overlapping
patches with any padding placed on the oldest side and incorporates their tokens
sequentially through a recurrent scan.

\paragraph{Recurrent Depth.}
Recent language models reason through continuous latent representations
\citep{coconut} or iteratively refine hidden states through recurrent depth
\citep{recurrentdepth}. MARO draws on shared recurrent computation but assigns
a distinct historical role to each step.
Each recurrence introduces one additional older patch instead of processing
an unchanged token set again.

Recurrent processing is also well established in time series forecasting.
SegRNN passes temporal segments through a GRU in chronological order
\citep{lin2023segrnn}. xLSTMTime employs xLSTM cells
\citep{alharthi2024xlstmtime}. S-Mamba performs scans along either the
temporal or variable dimension \citep{wang2025mamba}. MARO differs from these
methods by starting at the forecast boundary and retaining the most recent patch
as a persistent anchor. Each subsequent step introduces one older patch. The
recurrence therefore encodes recency directly through the traversal order.

%% file: sections/method.tex
\section{Method}
\label{sec:method}

\subsection{Problem Setup and Overview}
\label{sec:preliminaries}
\label{sec:overview}

Let $\mathbf X\in\mathbb R^{C\times L}$ be the $C$-channel look-back window
and $\mathbf Y\in\mathbb R^{C\times H}$ the target. Calendar features are
observed at every history and horizon step and have dataset-specific width.

\begin{figure}[t]
   \centering
   \definecolor{chanA}{RGB}{99,156,99}
   \definecolor{chanB}{RGB}{178,112,122}
   \definecolor{chanC}{RGB}{104,120,181}
   \definecolor{creamfill}{RGB}{250,238,213}
   \definecolor{creamborder}{RGB}{193,148,72}
   \definecolor{bluefill}{RGB}{223,232,247}
   \definecolor{blueborder}{RGB}{88,111,163}
   \definecolor{greenfill}{RGB}{223,240,226}
   \definecolor{greenborder}{RGB}{86,140,96}
   \definecolor{anchfill}{RGB}{250,190,110}
   \definecolor{anchborder}{RGB}{193,120,30}
   \resizebox{\linewidth}{!}{%
   \begin{tikzpicture}[
     x=1cm,y=1cm,
     font=\sffamily\large,
     >=Latex,
     every node/.style={inner sep=1pt},
     flow/.style={-{Latex[length=2.6mm,width=2.4mm]},line width=1.5pt,draw=black!70},
     auxflow/.style={-{Latex[length=2.6mm,width=2.4mm]},line width=1.5pt,draw=creamborder},
     op/.style={draw=creamborder,fill=creamfill,rounded corners=3pt,line width=1.2pt,
                align=center,minimum height=12mm,inner sep=5pt},
     stage/.style={font=\sffamily\large\bfseries,text=black!55},
     note/.style={font=\sffamily\large,text=black!60,align=center},
     plus/.style={circle,draw=black!60,fill=white,line width=1.2pt,minimum size=8mm,inner sep=0pt},
   ]

   \foreach \y/\col in {0.95/chanA, 0/chanB, -0.95/chanC} {
     \begin{scope}[shift={(1.3,\y)}]
       \fill[\col!18] (0.55,-0.4) rectangle (1.3,0.4);
       \draw[black!45,line width=0.8pt] (-1.3,-0.4) rectangle (1.3,0.4);
       \draw[\col,line width=1.3pt] plot [smooth] coordinates
         {(-1.15,-0.08) (-0.7,0.24) (-0.25,-0.14) (0.2,0.2) (0.6,-0.18) (1.0,0.1) (1.2,0.26)};
     \end{scope}
   }
   \node[note,anchor=north] at (1.3,-1.5) {$\mathbf X\in\mathbb R^{C\times L}$};
   \foreach \y in {0.95,0,-0.95} {\draw[line width=1.8pt,draw=black!70] (2.6,\y) -- (3.15,\y);}
   \draw[line width=1.8pt,draw=black!70] (3.15,0.95) -- (3.15,-0.95);

   \node[op,minimum width=18mm] (revin) at (4.5,0) {RevIN};
   \draw[flow] (3.15,0) -- (revin.west);
   \node[note,anchor=north] at (5.0,-0.75) {$\mu_c,\varsigma_c$ stored};

   \foreach \i in {0,1,2} {
     \pgfmathsetmacro{\xx}{5.9+\i*0.64}
     \draw[black!55,fill=black!10,rounded corners=1.5pt,line width=1pt]
       (\xx,-0.45) rectangle ++(0.56,0.9);
   }
   \draw[anchborder,fill=anchfill,rounded corners=1.5pt,line width=1.5pt]
     (7.82,-0.45) rectangle ++(0.56,0.9);
   \draw[flow] (revin.east) -- (5.82,0);
   \node[note,anchor=south] at (7.15,0.55) {overlapping patches};
   \node[note,anchor=north,text=anchborder!75!black] at (8.1,-0.55) {$U_N$};

   \node[op,minimum width=24mm] (intra)  at (10.9,0) {Intra-Patch\\MLP};
   \node[plus]                  (p1)     at (13.4,0) {$+$};
   \node[op,minimum width=13mm] (proj)   at (15.3,0) {$W_P$};
   \node[op,minimum width=21mm] (tokmlp) at (17.9,0) {Token\\MLP};
   \node[plus]                  (p2)     at (20.2,0) {$+$};
   \draw[flow] (8.85,0) -- (intra.west);
   \draw[flow] (intra.east) -- (p1.west);
   \draw[flow] (p1.east) -- (proj.west);
   \draw[flow] (proj.east) -- (tokmlp.west);
   \draw[flow] (tokmlp.east) -- (p2.west);
   \draw[flow,rounded corners=4pt] (9.3,0) -- (9.3,1.15) -- (13.4,1.15) -- (p1.north);
   \draw[flow,rounded corners=4pt] (16.4,0) -- (16.4,1.15) -- (20.2,1.15) -- (p2.north);
   \coordinate (petop) at (14.0,1.35);
   \begin{scope}[on background layer]
     \node[draw=black!40,dashed,rounded corners=5pt,line width=1.1pt,fill=black!4,
           fit=(intra)(p2)(petop),inner sep=8pt] {};
   \end{scope}
   \node[stage,fill=white,inner sep=3pt,anchor=west] at (10.1,1.63) {PATCH EMBEDDING};

   \node[op,minimum width=34mm] (chmix) at (23.4,0) {Sparse Channel Mixer\\+ Calendar};
   \draw[flow] (p2.east) -- (chmix.west);

   \begin{scope}[on background layer]
     \node[draw=blueborder,fill=bluefill!60,rounded corners=6pt,line width=1.4pt,
           minimum width=84mm,minimum height=46mm] (card) at (4.8,-5.15) {};
   \end{scope}
   \node[stage,text=blueborder,anchor=north] at (4.8,-3.23)
     {RECURRENT SCAN $U_{N-1}\!\to\!U_1$};

   \node[op,fill=anchfill!45,draw=anchborder,minimum width=26mm,minimum height=13mm]
     (init) at (2.4,-4.9) {Init from\\anchor $U_N$};
   \node[op,fill=white,draw=blueborder,minimum width=38mm,minimum height=13mm]
     (step) at (6.7,-4.9) {Inter-Patch Mixer\\+ Latent Update};
   \draw[flow] (init.east) -- (step.west);
   \draw[flow,rounded corners=4pt] (8.3,-4.25) -- (8.3,-3.68) -- (5.1,-3.68) -- (5.1,-4.25);
   \node[note,text=blueborder] at (6.7,-3.95) {$\times(N\!-\!1)$};

   \node[op,minimum width=72mm,minimum height=8mm] (cond) at (4.8,-6.75)
     {every step with anchor $a=U_N$ and calendar $\phi$};
   \draw[auxflow] (2.4,-6.35) -- (2.4,-5.55);
   \draw[auxflow] (6.7,-6.35) -- (6.7,-5.55);

   \draw[flow,rounded corners=6pt]
     (chmix.south) -- (23.4,-2.35) -- (-0.5,-2.35) -- (-0.5,-4.9) -- (init.west);

   \node[op,draw=blueborder,fill=bluefill,line width=1.8pt,minimum width=26mm,minimum height=13mm]
     (refine) at (11.3,-4.9) {Post-Scan\\Refine $\times R$};
   \draw[flow] (step.east) -- (refine.west);
   \node[note,anchor=north] at (11.3,-5.65) {no new patch};

   \node[op,draw=greenborder,fill=greenfill,minimum width=26mm,minimum height=13mm]
     (fusion) at (14.7,-4.9) {Checkpoint\\Fusion};
   \draw[flow] (refine.east) -- (fusion.west);
   \draw[flow,rounded corners=5pt] (9.0,-3.75) -- (14.7,-3.75) -- (fusion.north);
   \node[note,anchor=south] at (12.1,-3.67) {checkpoints $z^{(k-1)}$ for $k\in D$};

   \node[op,minimum width=22mm,minimum height=13mm] (dec) at (18.0,-4.9) {Decoder\\MLP};
   \draw[flow] (fusion.east)
     -- node[above,inner sep=3pt,font=\sffamily\LARGE] {$\bar z$} (dec.west);
   \node[plus] (add) at (20.4,-4.9) {$+$};
   \draw[flow] (dec.east) -- (add.west);
   \node[op,minimum width=32mm,minimum height=11mm] (resid) at (20.4,-3.1)
     {Raw + Calendar\\Residuals};
   \draw[auxflow] (resid.south) -- (add.north);
   \node[op,minimum width=20mm,minimum height=13mm] (deno) at (22.8,-4.9) {RevIN$^{-1}$};
   \draw[flow] (add.east) -- (deno.west);

   \begin{scope}[shift={(25.3,-4.9)}]
     \draw[black!45,line width=0.8pt] (-0.8,-0.6) rectangle (0.8,0.6);
     \draw[chanA,line width=1.3pt] plot [smooth] coordinates {(-0.65,0.28) (-0.2,0.44) (0.25,0.2) (0.65,0.38)};
     \draw[chanB,line width=1.3pt] plot [smooth] coordinates {(-0.65,-0.04) (-0.2,0.1) (0.25,-0.14) (0.65,0.05)};
     \draw[chanC,line width=1.3pt] plot [smooth] coordinates {(-0.65,-0.38) (-0.2,-0.2) (0.25,-0.44) (0.65,-0.26)};
   \end{scope}
   \draw[flow] (deno.east) -- (24.45,-4.9);
   \node[note,anchor=north] at (25.3,-5.65) {$\widehat{\mathbf Y}\in\mathbb R^{C\times H}$};

   \end{tikzpicture}%
   }
   \caption{Overview of the MARO architecture.}
   \label{fig:anchor_architecture}
   \end{figure}
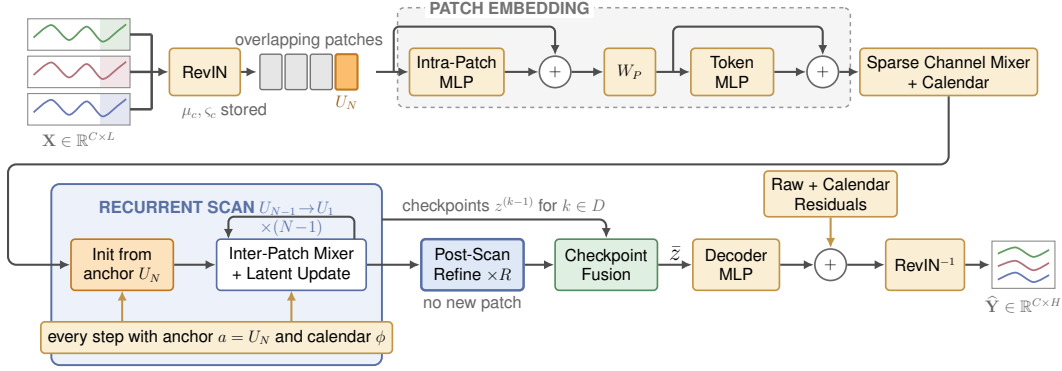

Figure~\ref{fig:anchor_architecture} illustrates the overall MARO architecture.
The normalized window is right-aligned and embedded into per-channel patch
tokens $U_{c,1},\dots,U_{c,N}$. Token $U_{c,N}$ touches the forecast boundary
(Section~\ref{sec:patching}). A sparse mixer exchanges information across
channels and incorporates calendar features within each patch
(Section~\ref{sec:channelmixer}). The recurrent module then operates on each
channel independently with shared weights. We suppress $c$ and write its
tokens as $U_1,\dots,U_N$. It initializes from $U_N$ and folds in
$U_{N-1},\dots,U_1$ while maintaining a temporal context and a forecast latent
(Section~\ref{sec:recurrence}). Exponentially spaced latent states are fused
into $\bar z$ (Section~\ref{sec:fusion}). An MLP decodes the fused state and
adds raw-input and future-calendar residuals (Section~\ref{sec:decoding}).

\subsection{Overlapping Patching}
\label{sec:patching}

MARO first applies reversible instance normalization (RevIN) to the look-back
window \citep{revin}. We denote the normalized window by
$\widetilde{\mathbf X}\in\mathbb R^{C\times L}$. It is split into
$N=\lceil\max(0,L-P)/S\rceil+1$ patches of length $P$ and stride $S$. Padding
is placed on the oldest side so every channel's most recent patch remains fully
observed.

Each patch passes through a residual sample-mixing MLP and a linear projection
to width $d$. A second residual MLP produces the final token
$U_{c,n}\in\mathbb R^d$.

\subsection{Sparse Channel Mixing and Calendar Conditioning}
\label{sec:channelmixer}
\label{sec:calendar}

For each target channel $c$, MARO scores the other channels using learned
channel embeddings, retains the $\kappa$ highest-scoring neighbours, and mixes
their patch tokens with the target token using normalized weights. A bounded
gate, initialized near zero, controls the resulting residual correction.
Setting $\kappa=0$ disables cross-channel exchange, whereas $\kappa=C-1$
recovers dense mixing.

The scorer is shared across patches and depends on learned channel identities,
not on the current input window. One sparse neighbourhood is therefore used
for every sample and patch, avoiding the per-sample $C\times C$ attention
tensor used by variate attention \citep{itransformer}. The selected neighbour
indices are discrete, but their scores retain gradients, so the routing can
change during training. We interpret the learned edges as routing choices, not
as recovered correlation structure. The complete scoring and gated-update
formulation is given in Appendix~\ref{app:channel-mixer}. Topology controls are
reported in Appendix~\ref{app:graph}.

Calendar conditioning enters at the same stage. For each historical patch, an
MLP encodes the calendar features at its two boundaries and midpoint, and the
resulting calendar token is added identically to every channel token. Because
the mixing weights sum to one, this shared token appears in both the weighted
neighbourhood token and the target token and therefore cancels from the
residual difference used for channel correction. Calendar information remains
in the token passed to the temporal scan without altering the sparse routing
correction.

A second MLP encodes the future calendar as tokens $f_t\in\mathbb R^d$. Its
boundary and midpoint tokens are compressed into the fixed future-context
vector $\phi\in\mathbb R^d$ used by every recurrent update. The full sequence
$\{f_t\}_{t=1}^{H}$ is retained for the calendar residual in
Section~\ref{sec:decoding}. After this stage, recurrence is channel-wise with
weights shared across channels. The neighbourhood fraction
$\delta=\kappa/(C-1)$ is reported for every setting in
Appendix~\ref{app:selected}.

\subsection{Recurrent Latent Refinement}
\label{sec:recurrence}

\begin{figure}[t]
   \centering
   \definecolor{creamfill}{RGB}{250,238,213}
   \definecolor{creamborder}{RGB}{193,148,72}
   \definecolor{bluefill}{RGB}{223,232,247}
   \definecolor{blueborder}{RGB}{88,111,163}
   \definecolor{tok}{RGB}{140,150,165}
   \definecolor{anchfill}{RGB}{250,190,110}
   \definecolor{anchborder}{RGB}{193,120,30}
   \definecolor{fusefill}{RGB}{223,240,226}
   \definecolor{fuseborder}{RGB}{86,140,96}
   \resizebox{\linewidth}{!}{%
   \begin{tikzpicture}[
     x=1cm,y=1cm,
     font=\sffamily\large,
     >=Latex,
     every node/.style={inner sep=1pt},
     flow/.style={-{Latex[length=2.4mm,width=2.2mm]},line width=1.4pt,draw=black!65},
     auxflow/.style={-{Latex[length=2.4mm,width=2.2mm]},line width=1.4pt,draw=creamborder},
     ckptflow/.style={-{Latex[length=2.4mm,width=2.2mm]},line width=1.5pt,draw=blueborder},
     op/.style={draw=creamborder,fill=creamfill,rounded corners=3pt,line width=1.1pt,align=center,minimum height=10mm,inner sep=5pt},
     ptok/.style={draw=tok!70!black,fill=tok!35,rounded corners=2pt,line width=1pt,minimum width=15mm,minimum height=9mm},
     atok/.style={ptok,fill=anchfill,draw=anchborder,line width=1.5pt},
     st/.style={circle,draw=blueborder,fill=bluefill,line width=1.2pt,minimum size=14mm,inner sep=0pt},
     ck/.style={st,line width=3pt},
     dlab/.style={font=\sffamily\large\bfseries,text=blueborder,anchor=south},
   ]

   \node[atok] (p0) at (0,3.1) {$U_N$};
   \foreach \s/\lbl in {1/{$U_{N-1}$},2/{$U_{N-2}$},3/{$U_{N-3}$},4/{$U_{N-4}$},6/{$U_{2}$},7/{$U_{1}$}} {
     \node[ptok] (p\s) at (\s*2.3,3.1) {\lbl};
   }
   \node[font=\sffamily\LARGE,text=black!45] at (11.5,3.1) {$\cdots$};

   \foreach \s/\lbl in {0/{$z^{(0)}$},1/{$z^{(1)}$},2/{$z^{(2)}$},3/{$z^{(3)}$},
                        4/{$z^{(4)}$},6/{$z^{(N-2)}$},7/{$z^{(N-1)}$}} {
     \node[st] (z\s) at (\s*2.3,1.2) {\lbl};
   }
   \node[ck] (z5) at (11.5,1.2) {$z^{(2^m\!-\!1)}$};
   \foreach \s/\lbl in {0/{$z^{(0)}$},1/{$z^{(1)}$},3/{$z^{(3)}$}} {
     \node[ck] at (\s*2.3,1.2) {\lbl};
   }
   \foreach \s/\k in {0/1,1/2,3/4} {
     \node[dlab,anchor=south west] at (\s*2.3+0.3,1.95) {$k=\k$};
   }
   \node[dlab,anchor=south] at (11.5,1.95) {$k=2^m$};

   \foreach \s in {0,1,2,3,4,6,7} {
     \draw[flow] (p\s.south) -- (z\s.north);
   }
   \foreach \s/\t in {0/1,1/2,2/3,3/4,6/7} {
     \draw[flow] (z\s.east) -- (z\t.west);
   }
   \draw[flow] (z4.east) -- (z5.west);
   \draw[flow] (z5.east) -- (z6.west);

   \node[op,fill=bluefill,draw=blueborder,line width=3pt,minimum width=28mm]
     (extra) at (19.3,1.2) {Refine $\times R$};
   \draw[flow] (z7.east) -- (extra.west);
   \node[dlab] at (19.3,1.95) {$k=N$};

   \draw[creamborder,line width=1.6pt] (0,-0.6) -- (18.4,-0.6);
   \foreach \x in {0,2.3,4.6,6.9,9.2,11.5,13.8,16.1}
     {\draw[auxflow] (\x,-0.6) -- (\x,0.5);}
   \draw[auxflow] (18.4,-0.6) -- (18.4,0.7);

   \node[draw=fuseborder,fill=fusefill,rounded corners=4pt,line width=1.5pt,
         minimum width=224mm,minimum height=9mm,align=center,
         font=\sffamily\large\bfseries,text=fuseborder!55!black]
     (fuse) at (9.55,-1.75) {Checkpoint Fusion};

   \foreach \s in {0,1,3,5} {
     \draw[blueborder,line width=1.6pt] (z\s.200) -- (\s*2.3-1.15,0.961);
     \draw[ckptflow] (\s*2.3-1.15,0.961) -- (\s*2.3-1.15,-1.3);
   }
   \draw[blueborder,line width=1.6pt] (19.3,0.7) -- (20.2,0.7);
   \draw[ckptflow] (20.2,0.7) -- (20.2,-1.3);

   \draw[flow] (fuse.south) -- (9.55,-2.95);
   \node[anchor=north,font=\sffamily\Huge] at (9.55,-3.05) {$\bar z$};

   \end{tikzpicture}%
   }
   \caption{Recurrent processing from the most recent patch to the oldest.
   The anchor and future context condition every update. Selected intermediate
   states and the refined final state are fused into $\bar z$.}
   \label{fig:anchor_unrolled}
   \end{figure}
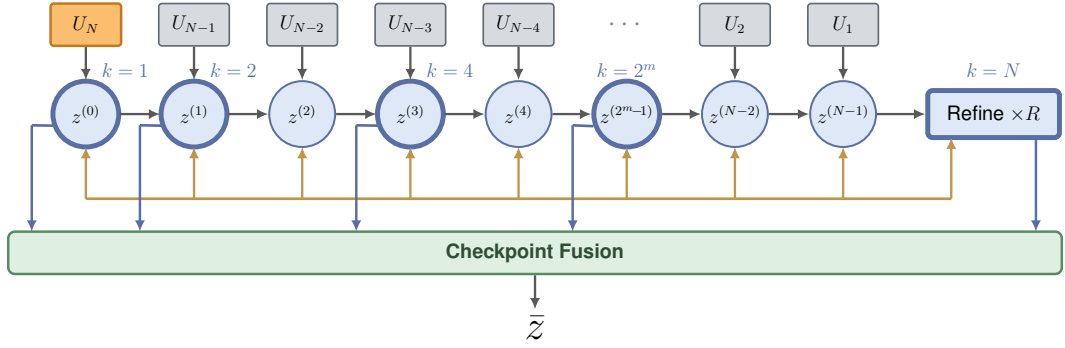

Figure~\ref{fig:anchor_unrolled} shows the shared recurrent module called the
\emph{scan}. It starts from $U_N$ and processes each older patch once in the
order $U_{N-1},\dots,U_1$. The scan runs independently for each channel. We
suppress the channel index and take all vectors below to lie in $\mathbb R^d$.

\paragraph{Context and latent state.}
At scan step $s$, the context state $h^{(s)}$ summarizes the patches
$U_N,\ldots,U_{N-s}$ incorporated so far without using horizon information.
The corresponding forecast latent $z^{(s)}$ is updated from that context, the
recent anchor, and the future calendar. Separating the two states keeps temporal
comparison in $h^{(s)}$ distinct from horizon-conditioned refinement in
$z^{(s)}$.

\paragraph{Initialization.}
The fixed anchor $a=U_N$ initializes both states.
\begin{equation}
\label{eq:init}
h^{(0)}=a,
\qquad
z^{(0)}=a+\mathrm{MLP}_{\mathrm{init}}\big([\,a\,\Vert\,\phi\,]\big),
\end{equation}
where $\Vert$ denotes feature concatenation. The latent begins at the forecast
boundary after adjustment for the future calendar.

\paragraph{Positional embedding.}
Step $s=1,\dots,N-1$ consumes $U_{N-s}$. Its normalized scan position is
encoded as
\begin{equation}
\label{eq:step}
\mathrm{pos}_s=\mathrm{MLP}_{\mathrm{step}}\left(
\frac{s}{N-1},\;\left(\frac{s}{N-1}\right)^2,\;
\sin\frac{2\pi s}{N-1},\;\cos\frac{2\pi s}{N-1}\right).
\end{equation}
This allows the same encoder to handle different scan lengths.

\paragraph{Progressive temporal mixer.}
The inter-patch MLP compares the incoming older patch with the accumulated
recent context.
\begin{equation}
\label{eq:relation}
\rho_s=\mathrm{MLP}_{\mathrm{rel}}\Big(\big[\,h^{(s-1)}\,\Vert\,U_{N-s}\,\Vert\,U_{N-s}\odot h^{(s-1)}\,\Vert\,\mathrm{pos}_s\,\big]\Big).
\end{equation}
The elementwise product $U_{N-s}\odot h^{(s-1)}$ is a multiplicative
interaction between the incoming patch and the accumulated context. Reusing
the same mixer at every step extends this pairwise comparison over the complete
temporal window without adding parameters with scan depth.

\paragraph{Gated additive updates.}
A shared bounded gate $g_r=\bar g_r\sigma(\ell_r)$ controls both state updates.
\begin{align}
h^{(s)} &= h^{(s-1)} + g_r\,\rho_s, \label{eq:ctx-update}\\
z^{(s)} &= z^{(s-1)} + g_r\,\underbrace{\mathrm{MLP}_{\mathrm{upd}}\Big(\big[\,z^{(s-1)}\,\Vert\,h^{(s)}\,\Vert\,\rho_s\,\Vert\,a\,\Vert\,\phi\,\Vert\,\mathrm{pos}_s\,\big]\Big)}_{u_s}. \label{eq:latent-update}
\end{align}
The gate is shared to regulate the update strength of both recurrent states
with one parameter. It is applied once to the context update and once to the
latent update. The latter uses the updated context together with the temporal
relation and fixed conditioning from the anchor and future calendar.

The relation $\rho_s$ reaches the latent through the accumulated context and a
direct input. These paths provide the temporal summary and the current patch
relation, respectively. The anchor $a$ and future context $\phi$ remain fixed
throughout the scan.

\subsection{Checkpointed Multi-Depth Fusion}
\label{sec:fusion}

MARO preserves states at several
temporal depths. Depth $k$ denotes $z^{(k-1)}$. This state has consumed the $k$
most recent patches. Checkpoints double in depth until the full window.
\begin{equation}
\label{eq:depths}
D=\{1,2,4,\dots\}\cup\{N\},\qquad |D|=O(\log N).
\end{equation}
This concentrates resolution near the forecast boundary while using only
$O(\log N)$ states. Shared softmax weights are initialized uniformly and combine
them without increasing the latent width.
\begin{equation}
\label{eq:fusion}
\alpha=\mathrm{softmax}\big(\ell^{\mathrm{fuse}}\big)\in\mathbb R^{|D|},
\qquad
\bar z_c=\sum_{k\in D}\alpha_k\,z^{(k-1)}_c
\quad\text{for every channel } c.
\end{equation}
$\bar z_c\in\mathbb R^d$ regardless of $N$. Sharing $\alpha$ gives one
dataset-level preference over temporal depth.

\paragraph{Induced recency prior.}

\begin{proposition}
\label{prop:recency}
The purely additive latent path with a step-independent gate makes checkpoint
fusion exactly a learned monotone recency prior.
\begin{equation}
\label{eq:lambda}
\bar z=z^{(0)}+g_r\sum_{s=1}^{N-1}\lambda_s\,u_s
+\alpha_N g_r\sum_{r=1}^{R}v_r,
\qquad
\lambda_s=\sum_{k\in D,\;k>s}\alpha_k ,
\end{equation}
where $v_r$ is the $r$-th optional post-scan increment of
Section~\ref{sec:refinement}. The weights $\lambda_s$ are non-increasing in
$s$. Their boundary values are $\lambda_1=1-\alpha_1$ and
$\lambda_{N-1}=\alpha_N$. The anchor coefficient is identically one.
\end{proposition}

Appendix~\ref{app:proof} provides the proof. No
older-patch increment receives more weight than a newer one. The post-scan
term refines the complete temporal summary without changing that ordering.

\subsection{Post-Scan Refinement and Forecast Decoding}
\label{sec:refinement}
\label{sec:decoding}

After the scan MARO applies the same latent-update MLP $R$ additional times to
the deepest state. Because these steps consume no new patch, the relation-input
slot occupied by $\rho_s$ during the scan is set to the zero vector
$\mathbf 0\in\mathbb R^d$.
\begin{equation}
\label{eq:refine}
z\ \leftarrow\ z+g_r\,\mathrm{MLP}_{\mathrm{upd}}\big([\,z\,\Vert\,h^{(N-1)}\,\Vert\,\mathbf 0\,\Vert\,a\,\Vert\,\phi\,\Vert\,\mathrm{pos}_{N-1}\,]\big).
\end{equation}
The refined state replaces the deepest checkpoint before fusion. The decoder
maps each fused state $\bar z_c$ to an $H$-step latent prediction
$\widehat{\mathbf Y}^{\mathrm{lat}}_{c,:}$. The prediction on the normalized
scale also includes zero-initialized residual branches from the normalized
input and future calendar.
\begin{equation}
\label{eq:decsum}
\widehat{\mathbf Y}^{\mathrm{norm}}
=\widehat{\mathbf Y}^{\mathrm{raw}}
+\widehat{\mathbf Y}^{\mathrm{lat}}
+\widehat{\mathbf Y}^{\mathrm{cal}},
\end{equation}
where $\widehat{\mathbf Y}^{\mathrm{raw}}$ is a learned linear projection from
$\widetilde{\mathbf X}$ to the $H$ forecast steps, and
$\widehat{Y}^{\mathrm{cal}}_{c,t}=v_c^\top f_t+b_c$ uses a learned
channel-specific vector $v_c$ and bias $b_c$. Inverse RevIN then restores the
original scale to obtain the forecast $\widehat{\mathbf Y}$.

%% file: sections/experiments.tex
\section{Experiments}
\label{sec:experiments}

\paragraph{Protocol.}
The long-term suite comprises ETTh1, ETTh2, ETTm1, and ETTm2
\citep{informer}, Weather \citep{weather}, Electricity \citep{electricity},
Traffic \citep{pems}, and Solar \citep{solar}. The short-term suite comprises
PEMS03, PEMS04, PEMS07, and PEMS08 \citep{pems}. The long-term horizons are
$H\in\{96,192,336,720\}$, and the PEMS horizons are
$H\in\{12,24,48,96\}$.

The long-term baselines are SRSNet \citep{srsnet}, TimeKAN \citep{timekan},
Amplifier \citep{amplifier}, iTransformer \citep{itransformer}, TimeMixer
\citep{timemixer}, and PatchTST \citep{patchtst}. The short-term baselines are
CycleNet/MLP \citep{cyclenet}, iTransformer \citep{itransformer}, SCINet
\citep{scinet}, and PatchTST \citep{patchtst}. The efficiency comparison also
includes DLinear \citep{dlinear}. Long-term baseline results are quoted from
the SRSNet paper \citep{srsnet}; short-term baseline results are quoted from
their published sources.

A fixed look-back length rewards whichever method that length happens to suit
\citep{channelindep}. Following SRSNet, we select the look-back
length for each long-term dataset and horizon from $\{96,336,512\}$. PEMS uses
a fixed 96-step look-back. MARO is trained with mean squared error. The best
value is bold and the second best is underlined. Appendix~\ref{app:setup}
describes the datasets and evaluation protocol. Appendix~\ref{app:hparams}
gives the search settings, and Appendix~\ref{app:results} reports the full
long-term results.

\subsection{Main Results}
\label{sec:main-results}

\begingroup
\setlength{\intextsep}{6pt plus 1pt minus 1pt}

\input{tables/table3_avg_ett_collapsed}

\noindent\textbf{Long-term forecasting.}\quad
Table~\ref{tab:avg-ett-collapsed} averages the four horizons of each dataset
and merges the four ETT subsets into one row. MARO achieves the best or tied
best result on four of the five groups. The largest gains occur on Traffic and
Electricity, where MARO improves over the strongest baseline by about five and
three percent in MSE, respectively. On the combined ETT group, it lowers MSE
from 0.335 to 0.329 while retaining the best MAE. On Weather, MARO ties the
best MSE and attains the lowest MAE. Solar is the only group it does not lead.
Its MSE and MAE trail SRSNet by 0.003 and 0.007.
Appendix~\ref{app:results} provides the disaggregated results for every dataset
and horizon.

\input{tables/table_pems_compact}

\noindent\textbf{Short-term forecasting.}\quad
The PEMS comparison follows the baseline set and 96-step-input protocol
reported by CycleNet \citep{cyclenet}. MARO ranks first in 8 of 16 MSE cells
and 6 of 16 MAE cells. The next-highest counts are 4 and 5, respectively. It
leads both metrics at $H\in\{12,24\}$ on PEMS03, PEMS07, and PEMS08, and it
also retains the best MSE on PEMS08 at $H\in\{48,96\}$. PEMS04 is the clear
exception, where SCINet leads all four horizons. At longer horizons,
iTransformer and SCINet become more competitive.

\endgroup
\vspace{-6pt}

\subsection{Ablation Study}
\label{sec:maro-ablations}

Table~\ref{tab:maro-ablation-average} compares MARO with variants that remove
its calendar input and channel mixer as well as its temporal mixing stages. It
also includes an anchor-only predictor and a \emph{full-window MLP} that
replaces the recurrent scan and fusion with one MLP over all patch tokens.

The complete model performs best on nearly every dataset and metric. The most
consistent degradation appears when temporal processing is weakened. Decoding
only the anchor discards useful history while the full-window MLP loses the
progressive integration and multi-depth fusion of MARO. Removing refinement and
processing the full window at once also reduce accuracy across all datasets.
Channel mixing is particularly important
on ETTh2 but has a smaller effect on Weather and Electricity. Calendar
conditioning generally helps, while its effect on Weather is negligible. The
topology controls in Appendix~\ref{app:graph} separately evaluate whether the
selected channel graph must be learned.

\input{tables/table_ablation_four_horizon_average}
\FloatBarrier

\subsection{Recent-to-Old and Old-to-Recent Traversal}
\label{sec:maro-anchor-order}

\input{tables/table_anchor_order}

Table~\ref{tab:maro-anchor-order} compares the two complete traversal
conventions on ETTh1, ETTm1, and Weather
at the 96- and 336-step horizons. Each traversal processes every patch once.
Recent-to-old traversal obtains lower MSE in five of the six settings. The
exception is ETTh1 at $H=336$, where old-to-recent traversal is lower by
0.0008.

\paragraph{Computational cost.}
Figure~\ref{fig:efficiency-bubble} compares forward FLOPs, inference latency,
and peak training memory for MARO and four baselines. Exact measurements,
including parameter counts and first-epoch training time, are reported in
Table~\ref{tab:maro-efficiency} in Appendix~\ref{app:efficiency}. All
measurements use a 512-step look-back, a 96-step forecast horizon, and a single
RTX 4060 Laptop GPU. For Electricity, MARO
completes each epoch $12.15\times$ faster than PatchTST and $12.82\times$ faster
than TimeKAN. It also uses roughly one sixth as many parameters as PatchTST and
about one hundredth of its FLOPs. PatchTST processes each channel independently
through its Transformer layers, even though encoder parameters can be shared
across channels. MARO instead uses a lighter recurrent scan with weights shared
across channels.
On Weather, only DLinear requires fewer FLOPs than MARO. MARO's epoch time,
latency, and peak memory lie between those of iTransformer and TimeKAN and
remain far below PatchTST at its wide-patch (192) operating point, which uses
roughly $390\times$ more FLOPs and takes $11\times$ longer per epoch. ETTh1 is the
exception. On ETTh1, the sequential scan under-utilizes the GPU because the
dataset has only seven channels, making training slower than DLinear,
iTransformer, and PatchTST despite similar or lower FLOPs. This overhead
becomes smaller as the number of channels increases.
\input{sections/fig_efficiency_bubble}
\FloatBarrier

%% file: tables/table3_avg_ett_collapsed.tex
\begin{table}[t]
\centering
\caption{Long-term results averaged over
$H\in\{96,192,336,720\}$, with the four ETT subsets merged into one group.
Full per-dataset, per-horizon results and standard deviations are reported in
Appendix~\ref{app:results}.}
\label{tab:avg-ett-collapsed}
\scriptsize
\setlength{\tabcolsep}{3pt}
\begin{tabular*}{\textwidth}{@{\extracolsep{\fill}}lcccccccccccccc@{}}
\toprule
 & \multicolumn{2}{c}{MARO (Ours)} & \multicolumn{2}{c}{SRSNet} & \multicolumn{2}{c}{TimeKAN} & \multicolumn{2}{c}{Amplifier} & \multicolumn{2}{c}{iTransformer} & \multicolumn{2}{c}{TimeMixer} & \multicolumn{2}{c}{PatchTST} \\
\cmidrule(lr){2-3} \cmidrule(lr){4-5} \cmidrule(lr){6-7} \cmidrule(lr){8-9} \cmidrule(lr){10-11} \cmidrule(lr){12-13} \cmidrule(lr){14-15}
Dataset & MSE & MAE & MSE & MAE & MSE & MAE & MSE & MAE & MSE & MAE & MSE & MAE & MSE & MAE \\
\midrule
ETT (Avg) & \textbf{0.329} & \textbf{0.372} & \underline{0.335} & \underline{0.375} & 0.340 & 0.380 & 0.346 & 0.383 & 0.351 & 0.384 & 0.346 & 0.383 & 0.343 & 0.381 \\
Weather & \textbf{0.223} & \textbf{0.261} & \underline{0.225} & 0.265 & 0.226 & 0.269 & \textbf{0.223} & 0.263 & 0.230 & 0.269 & \underline{0.225} & \underline{0.262} & \textbf{0.223} & \underline{0.262} \\
Electricity & \textbf{0.155} & \textbf{0.252} & \underline{0.160} & \underline{0.254} & 0.163 & 0.257 & 0.162 & 0.255 & 0.162 & 0.258 & 0.184 & 0.281 & 0.170 & 0.270 \\
Traffic & \textbf{0.372} & \textbf{0.265} & \underline{0.391} & \underline{0.270} & 0.419 & 0.285 & 0.412 & 0.290 & 0.397 & 0.280 & 0.409 & 0.278 & 0.396 & 0.275 \\
Solar & \underline{0.186} & \underline{0.245} & \textbf{0.183} & \textbf{0.238} & 0.198 & 0.263 & 0.202 & 0.281 & 0.201 & 0.260 & 0.203 & 0.260 & 0.200 & 0.284 \\
\midrule
\multicolumn{1}{l}{\# Best / Total} & \textbf{4/5} & \textbf{4/5} & \underline{1/5} & \underline{1/5} & 0/5 & 0/5 & \underline{1/5} & 0/5 & 0/5 & 0/5 & 0/5 & 0/5 & \underline{1/5} & 0/5 \\
\bottomrule
\end{tabular*}
\end{table}

%% file: tables/table_pems_compact.tex
\begin{table}[H]
\centering
\caption{Full PEMS forecasting results. Baseline values are quoted from the
published CycleNet comparison.}
\label{tab:full-results-pems-datasets}
\scriptsize
\setlength{\tabcolsep}{2.8pt}
\begin{tabular*}{\textwidth}{@{\extracolsep{\fill}}ll|cc|cc|cc|cc|cc@{}}
\toprule
 & & \multicolumn{2}{c|}{MARO (Ours)}
 & \multicolumn{2}{c|}{CycleNet/MLP}
 & \multicolumn{2}{c|}{iTransformer}
 & \multicolumn{2}{c|}{SCINet}
 & \multicolumn{2}{c}{PatchTST} \\
Dataset & $H$ & MSE & MAE & MSE & MAE & MSE & MAE & MSE & MAE & MSE & MAE \\
\midrule
PEMS03 & 12 & \textbf{0.064} & \textbf{0.168} & \underline{0.066} & \underline{0.172} & 0.071 & 0.174 & \underline{0.066} & \underline{0.172} & 0.099 & 0.216 \\
 & 24 & \textbf{0.083} & \textbf{0.190} & 0.089 & 0.201 & 0.093 & 0.201 & \underline{0.085} & \underline{0.198} & 0.142 & 0.259 \\
 & 48 & 0.131 & 0.239 & 0.136 & 0.247 & \textbf{0.125} & \textbf{0.236} & \underline{0.127} & \underline{0.238} & 0.211 & 0.319 \\
 & 96 & 0.188 & 0.298 & 0.182 & \underline{0.282} & \textbf{0.164} & \textbf{0.275} & \underline{0.178} & 0.287 & 0.269 & 0.370 \\
\midrule
PEMS04 & 12 & \underline{0.075} & \underline{0.179} & 0.078 & 0.186 & 0.078 & 0.183 & \textbf{0.073} & \textbf{0.177} & 0.105 & 0.224 \\
 & 24 & \underline{0.094} & \underline{0.204} & 0.099 & 0.212 & 0.095 & 0.205 & \textbf{0.084} & \textbf{0.193} & 0.153 & 0.275 \\
 & 48 & 0.136 & 0.250 & 0.133 & 0.248 & \underline{0.120} & \underline{0.233} & \textbf{0.099} & \textbf{0.211} & 0.229 & 0.339 \\
 & 96 & 0.196 & 0.309 & 0.167 & 0.281 & \underline{0.150} & \underline{0.262} & \textbf{0.114} & \textbf{0.227} & 0.291 & 0.389 \\
\midrule
PEMS07 & 12 & \textbf{0.055} & \textbf{0.148} & \underline{0.062} & \underline{0.162} & 0.067 & 0.165 & 0.068 & 0.171 & 0.095 & 0.207 \\
 & 24 & \textbf{0.081} & \textbf{0.183} & \underline{0.086} & 0.192 & 0.088 & \underline{0.190} & 0.119 & 0.225 & 0.150 & 0.262 \\
 & 48 & \underline{0.128} & 0.238 & \underline{0.128} & \underline{0.234} & \textbf{0.110} & \textbf{0.215} & 0.149 & 0.237 & 0.253 & 0.340 \\
 & 96 & 0.148 & 0.257 & 0.176 & 0.268 & \textbf{0.139} & \underline{0.245} & \underline{0.141} & \textbf{0.234} & 0.346 & 0.404 \\
\midrule
PEMS08 & 12 & \textbf{0.075} & \textbf{0.176} & 0.082 & 0.185 & \underline{0.079} & \underline{0.182} & 0.087 & 0.184 & 0.168 & 0.232 \\
 & 24 & \textbf{0.100} & \textbf{0.197} & 0.117 & 0.226 & \underline{0.115} & \underline{0.219} & 0.122 & 0.221 & 0.224 & 0.281 \\
 & 48 & \textbf{0.152} & \underline{0.245} & \underline{0.169} & 0.268 & 0.186 & \textbf{0.235} & 0.189 & 0.270 & 0.321 & 0.354 \\
 & 96 & \textbf{0.202} & \underline{0.278} & 0.233 & 0.306 & \underline{0.221} & \textbf{0.267} & 0.236 & 0.300 & 0.408 & 0.417 \\
\midrule
\multicolumn{2}{l|}{\# Best / Total}
& \textbf{8/16} & \textbf{6/16} & 0/16 & 0/16
& \underline{4/16} & \underline{5/16}
& \underline{4/16} & \underline{5/16} & 0/16 & 0/16 \\
\bottomrule
\end{tabular*}
\end{table}

%% file: tables/table_ablation_four_horizon_average.tex
\begin{table}[H]
\centering
\caption{MARO component ablations averaged over $H\in\{96,192,336,720\}$.
Values are truncated to three decimal places.
The best value in each column is bold and the second best is underlined. The
full model is a repeated-seed mean, each ablation one validation-selected seed.}
\label{tab:maro-ablation-average}
\scriptsize
\setlength{\tabcolsep}{3pt}
\begin{tabular*}{\textwidth}{@{\extracolsep{\fill}}lcccccccc@{}}
\toprule
 & \multicolumn{2}{c}{ETTh1} & \multicolumn{2}{c}{ETTh2} & \multicolumn{2}{c}{Weather} & \multicolumn{2}{c}{Electricity} \\
\cmidrule(lr){2-3} \cmidrule(lr){4-5} \cmidrule(lr){6-7} \cmidrule(lr){8-9}
Variant & MSE & MAE & MSE & MAE & MSE & MAE & MSE & MAE \\
\midrule
Full MARO              & \textbf{0.396} & \textbf{0.423} & \textbf{0.316} & \textbf{0.373} & \underline{0.223} & \textbf{0.261} & \textbf{0.153} & \textbf{0.252} \\
\midrule
w/o calendar           & \underline{0.406} & 0.432 & 0.333 & 0.381 & \textbf{0.223} & 0.263 & 0.161 & \underline{0.255} \\
w/o channel mixing     & 0.408 & 0.433 & 0.371 & 0.408 & 0.224 & \underline{0.261} & \underline{0.157} & 0.256 \\
w/o inter-patch mixer  & 0.422 & 0.440 & \underline{0.332} & \underline{0.380} & 0.228 & 0.266 & 0.157 & \underline{0.255} \\
w/o intra-patch mixer  & 0.407 & \underline{0.431} & 0.337 & 0.384 & 0.228 & 0.266 & 0.168 & 0.263 \\
Anchor only (no scan)  & 0.416 & 0.438 & 0.337 & 0.385 & 0.232 & 0.267 & 0.163 & 0.261 \\
Full-window MLP        & 0.416 & 0.436 & 0.346 & 0.390 & 0.225 & 0.263 & 0.159 & 0.259 \\
w/o refinement         & 0.421 & 0.442 & 0.344 & 0.389 & 0.225 & 0.263 & 0.159 & 0.256 \\
\bottomrule
\end{tabular*}
\end{table}

%% file: tables/table_anchor_order.tex
\begin{table}[!htbp]
\centering
\caption{Test MSE of the two complete traversal conventions on ETTh1, ETTm1,
and Weather. Recent-to-old traversal anchors at the most recent patch and
scans toward the oldest. Old-to-recent traversal anchors at the oldest patch
and scans toward the most recent. Lower is better.}
\label{tab:maro-anchor-order}
\scriptsize
\setlength{\tabcolsep}{5pt}
\begin{tabular*}{\textwidth}{@{\extracolsep{\fill}}llcc@{}}
\toprule
Dataset & $H$ & Recent-to-old (MARO) & Old-to-recent \\
\midrule
ETTh1   & 96  & \textbf{0.3554} & 0.3816 \\
        & 336 & 0.3959          & \textbf{0.3951} \\
ETTm1   & 96  & \textbf{0.2842} & 0.3136 \\
        & 336 & \textbf{0.3599} & 0.3846 \\
Weather & 96  & \textbf{0.1460} & 0.1558 \\
        & 336 & \textbf{0.2442} & 0.2592 \\
\bottomrule
\end{tabular*}
\end{table}

%% file: sections/fig_efficiency_bubble.tex
\begin{figure}[!htbp]
\centering
\begin{tikzpicture}
\begin{groupplot}[
    group style={group size=3 by 1, horizontal sep=0.55cm},
    width=0.33\textwidth,
    height=0.27\textwidth,
    xmode=log,
    ymode=log,
    xmin=0.0005,
    xmax=300,
    ymin=0.0015,
    ymax=200,
    grid=both,
    minor grid style={draw=black!8},
    major grid style={draw=black!18},
    axis background/.style={fill=black!1},
    clip=true,
    clip marker paths=true,
    tick label style={font=\scriptsize},
    label style={font=\scriptsize},
    title style={font=\small\bfseries},
    xlabel={FLOPs/sample (G, log)},
    clip mode=individual,
]

\nextgroupplot[title={ETTh1},ylabel={Latency (ms/sample, log)}]
\addplot+[only marks,mark=*,mark size=1.30pt,draw=blue!70!black,fill=blue!55,fill opacity=0.72] coordinates {(0.001,0.004)};
\addplot+[only marks,mark=*,mark size=2.60pt,draw=orange!80!black,fill=orange!70,fill opacity=0.72] coordinates {(0.028,0.338)};
\addplot+[only marks,mark=*,mark size=1.35pt,draw=teal!70!black,fill=teal!60,fill opacity=0.72] coordinates {(0.021,0.024)};
\addplot+[only marks,mark=*,mark size=1.80pt,draw=red!75!black,fill=red!60,fill opacity=0.72] coordinates {(0.134,0.052)};
\addplot+[only marks,mark=*,mark size=1.85pt,draw=violet!80!black,fill=violet!70,fill opacity=0.90,line width=0.8pt] coordinates {(0.089,0.130)};

\nextgroupplot[title={Electricity}]
\addplot+[only marks,mark=*,mark size=1.30pt,draw=blue!70!black,fill=blue!55,fill opacity=0.72] coordinates {(0.063,0.044)};
\addplot+[only marks,mark=*,mark size=6.50pt,draw=orange!80!black,fill=orange!70,fill opacity=0.72] coordinates {(2.170,32.348)};
\addplot+[only marks,mark=*,mark size=2.20pt,draw=teal!70!black,fill=teal!60,fill opacity=0.72] coordinates {(3.918,1.368)};
\addplot+[only marks,mark=*,mark size=5.50pt,draw=red!75!black,fill=red!60,fill opacity=0.72] coordinates {(90.365,24.970)};
\addplot+[only marks,mark=*,mark size=1.75pt,draw=violet!80!black,fill=violet!70,fill opacity=0.90,line width=0.8pt] coordinates {(0.968,0.625)};

\nextgroupplot[title={Weather}]
\addplot+[only marks,mark=*,mark size=1.30pt,draw=blue!70!black,fill=blue!55,fill opacity=0.72] coordinates {(0.004,0.007)};
\addplot+[only marks,mark=*,mark size=4.10pt,draw=orange!80!black,fill=orange!70,fill opacity=0.72] coordinates {(0.173,2.348)};
\addplot+[only marks,mark=*,mark size=1.35pt,draw=teal!70!black,fill=teal!60,fill opacity=0.72] coordinates {(0.061,0.044)};
\addplot+[only marks,mark=*,mark size=6.10pt,draw=red!75!black,fill=red!60,fill opacity=0.72] coordinates {(17.418,3.438)};
\addplot+[only marks,mark=*,mark size=1.35pt,draw=violet!80!black,fill=violet!70,fill opacity=0.90,line width=0.8pt] coordinates {(0.045,0.182)};

\end{groupplot}

\matrix[draw=none,anchor=north,font=\scriptsize,column sep=5pt]
at ([yshift=-7mm]group c2r1.south) {
  \node[draw=blue!70!black,fill=blue!55,circle,minimum size=5pt,inner sep=0pt] {}; & \node {DLinear}; &
  \node[draw=orange!80!black,fill=orange!70,circle,minimum size=5pt,inner sep=0pt] {}; & \node {TimeKAN}; &
  \node[draw=teal!70!black,fill=teal!60,circle,minimum size=5pt,inner sep=0pt] {}; & \node {iTransformer}; &
  \node[draw=red!75!black,fill=red!60,circle,minimum size=5pt,inner sep=0pt] {}; & \node {PatchTST}; &
  \node[draw=violet!80!black,fill=violet!70,circle,minimum size=5pt,inner sep=0pt] {}; & \node {MARO}; \\
};
\end{tikzpicture}
\caption{Computational-efficiency comparison under the common protocol detailed
in Appendix~\ref{app:efficiency}. Both axes use logarithmic scales. Points
closer to the lower-left require fewer FLOPs and have lower inference latency.
Bubble size encodes peak training memory, and color identifies the model.}
\label{fig:efficiency-bubble}
\end{figure}
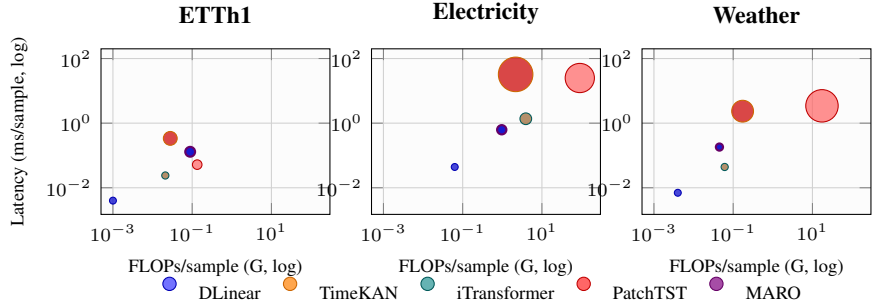

%% file: sections/conclusion.tex
\section{Conclusion}
\label{sec:conclusion}

We presented MARO, which anchors a forecast at the most recent patch and
recurrently incorporates older patches as context. Across long-term and
short-term benchmarks, MARO achieves strong overall accuracy, with its clearest
gains on Electricity and Traffic and at shorter PEMS horizons. Controlled
ablations identify the recent anchor and progressive integration of historical
context as the main contributors. Scan direction provides a smaller but
consistent benefit, while checkpoint fusion lets the forecast combine recent
and longer-range representations.

MARO also offers a favorable accuracy-efficiency trade-off. On the wider
Electricity and Weather benchmarks, it trains substantially faster and uses
far fewer FLOPs and less memory than PatchTST, and it is consistently faster
than TimeKAN in the measured settings. MARO uses more parameters than DLinear
and TimeKAN, and its sequential computation can be slower than lightweight
DLinear or iTransformer, particularly on small-channel data. Its benefit is
therefore not minimum latency alone, but stronger forecasting accuracy at a
moderate computational cost.

\paragraph{Limitations.}
The recurrent scan introduces a sequential dependency that can under-utilize
accelerators when the batch and channel dimensions are small. The learned
channel neighbourhood is fixed across input windows and therefore cannot adapt
to regime-specific dependencies. Learned routing also does not consistently
outperform fixed neighbourhoods.

\paragraph{Future work.}
Promising directions include parallel or chunked implementations of the
recent-to-old scan, input-adaptive sparse channel routing, and broader replicated
studies of anchor position and scan order. Extending MARO to irregularly
sampled series and distribution shifts would further test whether
recency-ordered computation remains effective beyond the current benchmarks.

%% file: sections/appendix.tex
\clearpage
\raggedbottom
\section{Experimental Setup}
\label{app:setup}

\subsection{Benchmarks and Evaluation Protocol}
\label{app:data}

Table~\ref{tab:maro-datasets} lists the twelve public benchmarks used in this
work. Eight form the long-term suite. The four ETT subsets contain seven load
and temperature variables recorded hourly or every 15 minutes
\citep{informer}. Weather contains 21 meteorological variables
\citep{weather}. Electricity records the hourly demand of 321 customers
\citep{electricity}. Traffic contains the hourly occupancy of 862 freeway
sensors \citep{pems}. Solar records the output of 137 photovoltaic plants
\citep{solar}. The four PEMS benchmarks contain five-minute traffic flow from
170 to 883 sensors \citep{pems}.

\input{tables/table_datasets}

We follow the chronological evaluation protocol introduced with Informer
\citep{informer}. Table~\ref{tab:maro-datasets} reports the sampling interval
and train-validation-test split of each benchmark. Standardization statistics
are fitted on the training segment. Sliding windows are constructed with
stride one from a look-back of length $L$ and forecast horizon $H$. MARO then
applies RevIN within each window as described in
Section~\ref{sec:patching}.

As an implementation detail, MARO augments the historical and future calendar
features with $\sin(2\pi t/W)$ and $\cos(2\pi t/W)$ before calendar encoding.
The dataset-specific cycle length $W$ is 24 for ETTh1/ETTh2, 96 for
ETTm1/ETTm2, 144 for Weather and Solar, and 168 for Electricity (ECL) and
Traffic.

\subsection{Error Measures}
\label{app:metrics}

Write $Y_{i,h,c}$ for a target and $\widehat Y_{i,h,c}$ for the corresponding
forecast. Index $i$ runs over the $n$ test windows. Index $h$ runs over the
horizon and $c$ runs over the channels. The reported metrics are
\begin{equation}
\label{eq:metrics}
\mathrm{MSE}=\frac{1}{nHC}\sum_{i=1}^{n}\sum_{h=1}^{H}\sum_{c=1}^{C}
\big(\widehat Y_{i,h,c}-Y_{i,h,c}\big)^2,
\qquad
\mathrm{MAE}=\frac{1}{nHC}\sum_{i=1}^{n}\sum_{h=1}^{H}\sum_{c=1}^{C}
\big|\widehat Y_{i,h,c}-Y_{i,h,c}\big| .
\end{equation}
MSE is used for training and model selection. Both metrics are averaged over
all channels and horizon steps. Lower values are better.

\subsection{Efficiency Measurements}
\label{app:efficiency}

Table~\ref{tab:maro-efficiency} reports the exact measurements visualized in
Figure~\ref{fig:efficiency-bubble}, together with parameter counts and
first-epoch training time.

\input{tables/table_efficiency}

\section{Hyperparameter Selection}
\label{app:hparams}

\subsection{Search Space and Budget}
\label{app:space}

We run 30 Optuna trials for each dataset and forecast horizon
\citep{optuna}. The long-term look-back length is selected from
$\{96,336,512\}$. Each selected configuration is evaluated with random seeds
3000 and 3002 as well as 3004. Table~\ref{tab:maro-space} gives the unified
search space.

\input{tables/table_searchspace}

\subsection{Selected Configurations}
\label{app:selected}

Tables~\ref{tab:maro-cfg-backbone} and~\ref{tab:maro-cfg-pems} report the best
backbone and optimization configurations for the long-term and PEMS
benchmarks, respectively. Table~\ref{tab:maro-cfg-gates} gives the remaining
long-term gate settings.

\input{tables/table_cfg_backbone}
\input{tables/table_cfg_pems}
\input{tables/table_cfg_gates}

\section{Full Results}
\label{app:results}
\label{app:full}

Tables~\ref{tab:full-results} and~\ref{tab:full-results-cont} report the full
long-term forecasting results.

\input{tables/table1_full}

The disaggregated ETT results show that MARO's advantage is less consistent on
ETTm1 and ETTm2 than on the hourly ETT datasets, although it remains close to
the best result. ETTm is sampled every 15 minutes, so a fixed step budget spans
a shorter real-time interval and contains fewer daily or weekly cycles. This
reduced long-period coverage may explain the smaller margins, particularly at
the longer horizons.

\section{Method Details and Mechanistic Controls}
\label{app:mech}

\subsection{Sparse Channel-Mixing Formulation}
\label{app:channel-mixer}

For completeness, this section gives the channel-mixing equations summarized
in Section~\ref{sec:channelmixer}. Each channel has a learned embedding
$e_c\in\mathbb R^{d_g}$. Two MLPs map it to a query
$q_c=\mathrm{MLP}_q(e_c)$ and a key $k_c=\mathrm{MLP}_k(e_c)$. The score from
channel $j$ to target channel $c$ is
\begin{equation}
\label{eq:graph}
G_{c,j}=\frac{q_c^{\top}k_j}{\sqrt{d_g}}\quad(c\neq j).
\end{equation}
The $\kappa$ highest-scoring channels form the neighbourhood $\mathcal N_c$.
A softmax over their scores and a learned self-score $\beta$ gives normalized
weights $w_{c,j}$. At patch $n$, the corresponding mixture is
\begin{equation}
A_{c,n}=w_{c,c}\,U_{c,n}+\sum_{j\in\mathcal N_c}w_{c,j}\,U_{j,n}.
\end{equation}
The original token receives a gated residual correction,
\begin{equation}
\label{eq:chanupdate}
U_{c,n}\ \leftarrow\ U_{c,n}
+g_{\mathrm{ch}}\cdot\mathrm{MLP}_v\big(A_{c,n}-U_{c,n}\big),
\end{equation}
where $g_{\mathrm{ch}}=\bar g_{\mathrm{ch}}
\sigma(\ell_{\mathrm{ch}})$ is a bounded learned gate initialized near zero.
The normalized weights sum to one, which is why a calendar token shared by all
channels cancels from $A_{c,n}-U_{c,n}$.

\subsection{Channel Topology}
\label{app:graph}

\input{tables/table_graph_ablation}

Table~\ref{tab:maro-graph-ablation} compares learned and fixed channel
neighbourhoods under the same MARO configuration. Each topology gives the
lowest mean MSE in two of the six settings. The benefit of learning the
channel graph therefore depends on the dataset rather than holding universally.

\subsection{Gate Parameterization and Context Gating}
\label{app:gatescreen}

\input{tables/table_gate_screen}

The gate alternatives produce only small changes from the bounded-gate
baseline. A separate context gate also gives no consistent improvement.
MARO therefore retains one shared recurrent gate.

\FloatBarrier
\section{Proof and Increment Analysis}
\label{app:proofs}

\subsection{Proof of Proposition~\ref{prop:recency}}
\label{app:proof}

Unrolling the additive latent update of \eqref{eq:latent-update} gives, for any
depth $k$,
\begin{equation}
\label{eq:unrolled}
z^{(k-1)}=z^{(0)}+g_r\sum_{s=1}^{k-1}u_s ,
\end{equation}
because $g_r$ does not depend on $s$ and the update contributes additively at
every step. Substituting \eqref{eq:unrolled} into the fusion
$\bar z=\sum_{k\in D}\alpha_k z^{(k-1)}$ and exchanging the two sums,
\begin{equation}
\bar z=\Big(\sum_{k\in D}\alpha_k\Big)z^{(0)}
+g_r\sum_{k\in D}\alpha_k\sum_{s=1}^{k-1}u_s
=z^{(0)}+g_r\sum_{s\ge 1}\Big(\sum_{k\in D,\;k>s}\alpha_k\Big)u_s ,
\end{equation}
where the anchor coefficient is $\sum_{k\in D}\alpha_k=1$ because $\alpha$ is a
softmax, and the inner sum collects every checkpoint deep enough to have
consumed step $s$. This is \eqref{eq:lambda} with
$\lambda_s=\sum_{k\in D,\,k>s}\alpha_k$.

Monotonicity is immediate, because $\lambda_s-\lambda_{s+1}=\alpha_{s+1}\ge 0$
when
$s+1\in D$ and $0$ otherwise, so $\lambda$ is non-increasing. For the boundary
values, $1\in D$ always, so
$\lambda_1=\sum_{k\in D,\,k>1}\alpha_k=1-\alpha_1$.  Also $N\in D$ is the unique
checkpoint exceeding $N-1$, so $\lambda_{N-1}=\alpha_N$. The post-scan
increments $v_r$ act only on the deepest state, which carries weight
$\alpha_N$, giving the final term of \eqref{eq:lambda}. \qed

We checked the identity numerically by fusing an explicitly simulated additive
scan and comparing against \eqref{eq:lambda} at
$N\in\{5,8,16,63,126\}$. The largest scan length among the selected
configurations is 126, attained by ETTm2 at $H=192$ and Solar at $H=720$
with $L=512$, $P=12$, and $S=4$. The two forms agree to at most
$8.9\times10^{-16}$ in absolute value, $\lambda$ is non-increasing at every
$N$, and both boundary values hold exactly.

\subsection{Increment Growth Along the Scan}
\label{app:increments}

Proposition~\ref{prop:recency} says the read-out receives
$\lambda_s\lVert g_r u_s\rVert$, and $\lambda_s$ is non-increasing by
construction. It does not constrain $\lVert g_r u_s\rVert$ itself. The
effective contribution can differ from $\lambda_s$ when update magnitudes
change across scan depth.

We instrument the ratio $\lVert g_r u_s\rVert/\lVert z^{(s-1)}\rVert$
at every scan step and average it over the batch and channels. On a 57-patch
ETTh1 configuration the profile is flat. It begins at $0.0126$ and ends at
$0.0111$. Its maximum is $1.02\times$ its first value across all 56 steps.

That single profile is not the whole picture. We later measured the same
quantity across the 24 baseline bounded-gate runs of the gate screen in
Appendix~\ref{app:mech}. Across these runs, 12 profiles are flat, nine grow
mildly and three grow. The increments therefore do not compound in most runs,
although they are not always flat. The monotone weighting of
Proposition~\ref{prop:recency} fully describes the recency prior when the profile
is flat. Mild growth makes the effective prior less steep than $\lambda_s$
alone implies. Separately, no
baseline gate trajectory saturated in those 24 runs. Neither observation
disturbs Proposition~\ref{prop:recency} itself, which is a statement about
non-negative checkpoint-fusion weights and not about the empirical size of
$\lVert g_r u_s\rVert$.  What they qualify is the empirical claim that an older
patch's contribution always diminishes in practice.

%% file: tables/table_datasets.tex
\begin{table}[H]
\centering
\caption{Benchmarks used in this work. $C$ is the number of variables and $T$ is the number of time steps. The final column gives the chronological train-validation-test split.}
\label{tab:maro-datasets}
\resizebox{\columnwidth}{!}{%
\begin{tabular}{llrrrl}
\toprule
Dataset & Domain & $C$ & $T$ & Sampling & Split (train/val/test) \\
\midrule
ETTh1 \citep{informer} & Transformer load & 7 & 17,420 & 1 hour & 8,640 / 2,880 / 2,880 \\
ETTh2 \citep{informer} & Transformer load & 7 & 17,420 & 1 hour & 8,640 / 2,880 / 2,880 \\
ETTm1 \citep{informer} & Transformer load & 7 & 69,680 & 15 min & 34,560 / 11,520 / 11,520 \\
ETTm2 \citep{informer} & Transformer load & 7 & 69,680 & 15 min & 34,560 / 11,520 / 11,520 \\
Weather \citep{weather} & Meteorology & 21 & 52,696 & 10 min & 36,887 / 5,270 / 10,539 \\
Electricity \citep{electricity} & Power demand & 321 & 26,304 & 1 hour & 18,412 / 2,632 / 5,260 \\
Traffic \citep{pems} & Road occupancy & 862 & 17,544 & 1 hour & 12,280 / 1,756 / 3,508 \\
Solar \citep{solar} & Photovoltaic output & 137 & 52,560 & 10 min & 36,792 / 5,256 / 10,512 \\
\midrule
PEMS03 \citep{pems} & Traffic flow & 358 & 26,208 & 5 min & 15,724 / 5,241 / 5,241 \\
PEMS04 \citep{pems} & Traffic flow & 307 & 16,992 & 5 min & 10,195 / 3,398 / 3,398 \\
PEMS07 \citep{pems} & Traffic flow & 883 & 28,224 & 5 min & 16,934 / 5,644 / 5,644 \\
PEMS08 \citep{pems} & Traffic flow & 170 & 17,856 & 5 min & 10,713 / 3,571 / 3,571 \\
\bottomrule
\end{tabular}%
}
\end{table}

%% file: tables/table_efficiency.tex
\begin{table}[H]
\centering
\caption{Measured cost of MARO and four representative baselines under a common protocol with $L=512$ and $H=96$ on one RTX 4060 Laptop GPU. Params counts trainable parameters, FLOPs is the forward cost of one sample and FLOPs$/C$ the same quantity per input channel, Epoch is the wall-clock time of the first training epoch, Latency is the evaluation-mode forward time per sample, and Memory is the peak allocation during training. ETTh1 uses batch size 32, Electricity uses batch size 4, and Weather uses batch size 16. Lower is better throughout.}
\label{tab:maro-efficiency}
\scriptsize
\setlength{\tabcolsep}{3pt}
\begin{tabular*}{\columnwidth}{@{\extracolsep{\fill}}llrrrrrr@{}}
\toprule
Dataset & Model & Params & FLOPs & FLOPs/$C$ & Epoch & Latency & Memory \\
 & & (M) & (G) & (G/ch.) & (s) & (ms/sample) & (MB) \\
\midrule
\multicolumn{8}{l}{\textit{ETTh1} ($C=7$, batch size 32)} \\
 & DLinear & 0.10 & 0.001 & 0.0001 & 0.62 & 0.004 & 21 \\
 & TimeKAN & 0.05 & 0.028 & 0.0040 & 7.54 & 0.338 & 423 \\
 & iTransformer & 0.95 & 0.021 & 0.0030 & 1.64 & 0.024 & 47 \\
 & PatchTST & 0.94 & 0.134 & 0.0191 & 1.91 & 0.052 & 162 \\
 & MARO (Ours) & 0.95 & 0.089 & 0.0128 & 4.87 & 0.130 & 167 \\
\midrule
\multicolumn{8}{l}{\textit{Electricity} ($C=321$, batch size 4)} \\
 & DLinear & 0.10 & 0.063 & 0.0002 & 12.10 & 0.044 & 37 \\
 & TimeKAN & 0.06 & 2.170 & 0.0068 & 1395.76 & 32.348 & 4180 \\
 & iTransformer & 5.05 & 3.918 & 0.0122 & 81.12 & 1.368 & 274 \\
 & PatchTST & 7.40 & 90.365 & 0.2815 & 1323.39 & 24.970 & 2729 \\
 & MARO (Ours) & 1.15 & 0.968 & 0.0030 & 108.88 & 0.625 & 146 \\
\midrule
\multicolumn{8}{l}{\textit{Weather} ($C=21$, batch size 16)} \\
 & DLinear & 0.10 & 0.004 & 0.0002 & 5.38 & 0.007 & 23 \\
 & TimeKAN & 0.06 & 0.173 & 0.0082 & 200.28 & 2.348 & 1408 \\
 & iTransformer & 1.21 & 0.061 & 0.0029 & 19.04 & 0.044 & 55 \\
 & PatchTST & 9.46 & 17.418 & 0.8294 & 486.50 & 3.438 & 3600 \\
 & MARO (Ours) & 0.42 & 0.045 & 0.0022 & 44.11 & 0.182 & 57 \\
\bottomrule
\end{tabular*}
\end{table}

%% file: tables/table_searchspace.tex
\begin{table}[H]
\centering
\caption{Unified Optuna search space for MARO.}
\label{tab:maro-space}
\resizebox{\columnwidth}{!}{%
\begin{tabular}{lll}
\toprule
Quantity & Symbol & Range or value \\
\midrule
Learning rate & $\eta$ & log-uniform, $[1\times10^{-5},5\times10^{-3}]$ \\
Look-back length & $L$ & $\{96,336,512\}$ \\
Token width & $d$ & $\{32,64,128,256,512\}$ \\
Inter-patch hidden width & $d_{\mathrm{ff}}$ & $\{64,128,256,512,1024\}$ \\
Latent MLP layers & $L_e$ & integer in $[1,4]$ \\
Intra-patch MLP layers & $L_d$ & integer in $[1,4]$ \\
Dropout & & uniform, $[0,0.40]$ \\
Patch length & $P$ & $\{12,24,32,48,64,96,112,128,144,192,256\}$, capped at $L$ \\
Stride ratio & $r$ & $\{0.30,0.50,0.75\}$ \\
Stride & $S$ & $\max(1,\mathrm{round}(rP))$, derived \\
Channel neighbourhood & $\kappa$ & $\{0,2,5,6,16,32,64,100,160,320\}$, capped at $C-1$ \\
Channel embedding width & $d_g$ & $\{4,8,16,32,64,128\}$, sampled only when $\kappa>0$ \\
Initial channel gate & $g_{\mathrm{ch}}$ & log-uniform, $[10^{-4},5\times10^{-2}]$, sampled only when $\kappa>0$ \\
Initial self score & $\beta$ & $\{0,0.5,1,2\}$, sampled only when $\kappa>0$ \\
Initial recurrent gate & $g_r$ & $\{0.01,0.025,0.05,0.10,0.20\}$ \\
Post-scan refinement steps & $R$ & $\{0,1,2,3\}$ \\
\midrule
Channel gate bound & $\bar g_{\mathrm{ch}}$ & fixed at $0.20$ \\
Recurrent gate bound & $\bar g_r$ & fixed at $0.30$ \\
Normalization & & fixed, RevIN with a learned affine map \\
Activation & & fixed, GELU \\
\bottomrule
\end{tabular}%
}
\end{table}

%% file: tables/table_cfg_backbone.tex
\begin{table}[H]
\centering
\caption{Best MARO backbone and optimization configurations for the long-term benchmarks. Here $\lambda$ denotes weight decay.}
\label{tab:maro-cfg-backbone}
\resizebox{\textwidth}{!}{%
\begin{tabular}{llrrrrrrrrrrrr}
\toprule
Dataset & $H$ & $L$ & $P$ & $S$ & $d$ & $d_{\mathrm{ff}}$ & $L_e$ & $L_d$ & $R$ & Dropout & $\eta$ & $\lambda$ & $B$ \\
\midrule
ETTh1 & 96 & 512 & 64 & 32 & 32 & 1024 & 1 & 1 & 1 & 0.014 & $2.645\times 10^{-4}$ & $1.0\times 10^{-5}$ & 256 \\
ETTh1 & 192 & 512 & 64 & 32 & 32 & 1024 & 1 & 4 & 0 & 0.082 & $3.677\times 10^{-4}$ & 0 & 256 \\
ETTh1 & 336 & 512 & 144 & 108 & 64 & 64 & 1 & 1 & 2 & 0.129 & $6.353\times 10^{-4}$ & $1.0\times 10^{-7}$ & 256 \\
ETTh1 & 720 & 336 & 112 & 84 & 128 & 128 & 1 & 1 & 1 & 0.267 & $1.812\times 10^{-3}$ & $1.0\times 10^{-3}$ & 256 \\
\midrule
ETTh2 & 96 & 512 & 32 & 16 & 32 & 512 & 4 & 1 & 3 & 0.320 & $4.370\times 10^{-4}$ & 0 & 256 \\
ETTh2 & 192 & 512 & 128 & 64 & 32 & 1024 & 1 & 1 & 3 & 0.166 & $6.428\times 10^{-4}$ & $1.0\times 10^{-6}$ & 256 \\
ETTh2 & 336 & 512 & 64 & 19 & 32 & 1024 & 1 & 1 & 0 & 0.009 & $1.510\times 10^{-4}$ & $1.0\times 10^{-3}$ & 256 \\
ETTh2 & 720 & 512 & 128 & 64 & 32 & 512 & 2 & 2 & 3 & 0.098 & $2.890\times 10^{-4}$ & 0 & 256 \\
\midrule
ETTm1 & 96 & 512 & 96 & 72 & 32 & 1024 & 1 & 1 & 0 & 0.385 & $1.110\times 10^{-3}$ & 0 & 256 \\
ETTm1 & 192 & 512 & 192 & 144 & 512 & 64 & 1 & 1 & 3 & 0.125 & $4.038\times 10^{-4}$ & $1.0\times 10^{-4}$ & 256 \\
ETTm1 & 336 & 512 & 96 & 48 & 32 & 512 & 2 & 2 & 1 & 0.077 & $2.498\times 10^{-4}$ & 0 & 256 \\
ETTm1 & 720 & 512 & 256 & 128 & 32 & 1024 & 1 & 1 & 1 & 0.081 & $1.402\times 10^{-4}$ & $1.0\times 10^{-5}$ & 256 \\
\midrule
ETTm2 & 96 & 336 & 24 & 7 & 64 & 512 & 1 & 1 & 0 & 0.359 & $2.150\times 10^{-4}$ & 0 & 256 \\
ETTm2 & 192 & 512 & 12 & 4 & 32 & 256 & 4 & 2 & 2 & 0.341 & $1.345\times 10^{-4}$ & 0 & 256 \\
ETTm2 & 336 & 512 & 48 & 24 & 32 & 64 & 1 & 1 & 3 & 0.305 & $6.552\times 10^{-4}$ & $1.0\times 10^{-3}$ & 256 \\
ETTm2 & 720 & 512 & 24 & 12 & 32 & 256 & 1 & 1 & 2 & 0.110 & $2.617\times 10^{-4}$ & $1.0\times 10^{-7}$ & 256 \\
\midrule
Weather & 96 & 512 & 192 & 96 & 32 & 128 & 3 & 3 & 0 & 0.192 & $4.722\times 10^{-4}$ & 0 & 64 \\
Weather & 192 & 512 & 144 & 72 & 64 & 1024 & 4 & 2 & 2 & 0.003 & $5.615\times 10^{-5}$ & 0 & 64 \\
Weather & 336 & 512 & 144 & 72 & 32 & 512 & 3 & 2 & 2 & 0.390 & $7.942\times 10^{-5}$ & 0 & 64 \\
Weather & 720 & 512 & 12 & 9 & 64 & 1024 & 3 & 4 & 2 & 0.153 & $2.385\times 10^{-5}$ & 0 & 64 \\
\midrule
Electricity & 96 & 512 & 256 & 192 & 128 & 128 & 3 & 1 & 0 & 0.152 & $6.634\times 10^{-4}$ & 0 & 16 \\
Electricity & 192 & 512 & 96 & 72 & 256 & 128 & 2 & 1 & 2 & 0.297 & $3.393\times 10^{-3}$ & 0 & 16 \\
Electricity & 336 & 512 & 256 & 128 & 256 & 1024 & 3 & 2 & 1 & 0.265 & $9.763\times 10^{-4}$ & 0 & 16 \\
Electricity & 720 & 512 & 96 & 72 & 32 & 1024 & 4 & 2 & 2 & 0.258 & $1.650\times 10^{-3}$ & 0 & 16 \\
\midrule
Traffic & 96 & 512 & 192 & 144 & 256 & 1024 & 2 & 3 & 2 & 0.004 & $8.215\times 10^{-4}$ & 0 & 16 \\
Traffic & 192 & 512 & 256 & 77 & 128 & 512 & 1 & 4 & 3 & 0.336 & $2.212\times 10^{-3}$ & 0 & 16 \\
Traffic & 336 & 512 & 144 & 108 & 256 & 1024 & 4 & 3 & 0 & 0.068 & $5.055\times 10^{-4}$ & 0 & 16 \\
Traffic & 720 & 512 & 192 & 96 & 256 & 1024 & 3 & 1 & 3 & 0.066 & $3.667\times 10^{-4}$ & 0 & 16 \\
\midrule
Solar & 96 & 512 & 96 & 48 & 256 & 256 & 1 & 3 & 3 & 0.309 & $1.387\times 10^{-3}$ & 0 & 64 \\
Solar & 192 & 512 & 96 & 48 & 64 & 256 & 3 & 2 & 2 & 0.067 & $4.518\times 10^{-4}$ & 0 & 16 \\
Solar & 336 & 512 & 12 & 6 & 512 & 256 & 1 & 3 & 0 & 0.243 & $7.979\times 10^{-5}$ & 0 & 16 \\
Solar & 720 & 512 & 12 & 4 & 32 & 256 & 1 & 4 & 3 & 0.233 & $6.316\times 10^{-4}$ & 0 & 16 \\
\bottomrule
\end{tabular}%
}
\end{table}

%% file: tables/table_cfg_pems.tex
\begin{table}[H]
\centering
\caption{Best MARO configurations for the PEMS benchmarks.}
\label{tab:maro-cfg-pems}
\resizebox{\textwidth}{!}{%
\begin{tabular}{llrrrrrrrrrrrrrrrr}
\toprule
Dataset & $H$ & $P$ & $S$ & $d$ & $d_{\mathrm{ff}}$ & $L_e$ & $L_d$ & $R$ & Dropout & $\eta$ & $\kappa$ & $\kappa/(C\!-\!1)$ & $d_g$ & $\beta$ & $g_{\mathrm{ch}}$ init & $g_r$ init & $B$ \\
\midrule
PEMS03 & 12 & 96 & 29 & 64 & 256 & 3 & 4 & 1 & 0.133 & $2.523\times 10^{-3}$ & 320 & 0.896 & 64 & 2.00 & $8.30\times 10^{-3}$ & 0.100 & 16 \\
PEMS03 & 24 & 12 & 6 & 64 & 512 & 1 & 3 & 0 & 0.109 & $4.786\times 10^{-3}$ & 320 & 0.896 & 4 & 0.00 & $5.01\times 10^{-3}$ & 0.100 & 16 \\
PEMS03 & 48 & 12 & 9 & 256 & 1024 & 1 & 3 & 0 & 0.109 & $8.417\times 10^{-4}$ & 6 & 0.017 & 128 & 2.00 & $3.02\times 10^{-3}$ & 0.200 & 16 \\
PEMS03 & 96 & 32 & 10 & 128 & 1024 & 1 & 3 & 0 & 0.185 & $2.426\times 10^{-3}$ & 100 & 0.280 & 64 & 2.00 & $4.75\times 10^{-3}$ & 0.010 & 16 \\
\midrule
PEMS04 & 12 & 12 & 6 & 512 & 1024 & 1 & 2 & 1 & 0.197 & $1.794\times 10^{-3}$ & 32 & 0.105 & 4 & 2.00 & $1.91\times 10^{-2}$ & 0.200 & 16 \\
PEMS04 & 24 & 48 & 36 & 128 & 1024 & 2 & 4 & 2 & 0.034 & $2.284\times 10^{-3}$ & 100 & 0.327 & 64 & 1.00 & $2.64\times 10^{-2}$ & 0.025 & 16 \\
PEMS04 & 48 & 24 & 7 & 64 & 1024 & 2 & 4 & 3 & 0.271 & $2.598\times 10^{-3}$ & 5 & 0.016 & 64 & 0.00 & $3.19\times 10^{-4}$ & 0.200 & 16 \\
PEMS04 & 96 & 48 & 14 & 512 & 256 & 3 & 3 & 1 & 0.133 & $1.441\times 10^{-3}$ & 32 & 0.105 & 8 & 0.00 & $4.74\times 10^{-2}$ & 0.050 & 16 \\
\midrule
PEMS07 & 12 & 12 & 6 & 128 & 512 & 1 & 1 & 3 & 0.022 & $3.094\times 10^{-3}$ & 16 & 0.018 & 32 & 0.50 & $2.28\times 10^{-2}$ & 0.200 & 8 \\
PEMS07 & 24 & 48 & 24 & 512 & 128 & 2 & 2 & 1 & 0.148 & $1.336\times 10^{-3}$ & 100 & 0.113 & 128 & 0.50 & $5.12\times 10^{-4}$ & 0.200 & 8 \\
PEMS07 & 48 & 48 & 24 & 512 & 512 & 4 & 2 & 1 & 0.295 & $3.891\times 10^{-4}$ & 16 & 0.018 & 16 & 0.50 & $1.47\times 10^{-2}$ & 0.050 & 8 \\
PEMS07 & 96 & 24 & 7 & 128 & 1024 & 1 & 1 & 2 & 0.249 & $1.082\times 10^{-3}$ & 320 & 0.363 & 32 & 0.50 & $3.05\times 10^{-4}$ & 0.025 & 8 \\
\midrule
PEMS08 & 12 & 48 & 14 & 256 & 1024 & 2 & 2 & 1 & 0.285 & $8.295\times 10^{-4}$ & 32 & 0.189 & 128 & 0.50 & $2.58\times 10^{-2}$ & 0.025 & 16 \\
PEMS08 & 24 & 12 & 4 & 256 & 1024 & 1 & 3 & 0 & 0.031 & $2.083\times 10^{-3}$ & 5 & 0.030 & 16 & 2.00 & $4.77\times 10^{-4}$ & 0.200 & 16 \\
PEMS08 & 48 & 8 & 4 & 32 & 512 & 2 & 2 & 2 & 0.068 & $2.332\times 10^{-3}$ & 64 & 0.379 & 128 & 2.00 & $7.33\times 10^{-4}$ & 0.100 & 16 \\
PEMS08 & 96 & 12 & 9 & 512 & 1024 & 2 & 2 & 2 & 0.167 & $1.342\times 10^{-3}$ & 2 & 0.012 & 32 & 0.00 & $4.69\times 10^{-2}$ & 0.200 & 16 \\
\bottomrule
\end{tabular}%
}
\end{table}

%% file: tables/table_cfg_gates.tex
\begin{table}[H]
\centering
\caption{Best MARO channel-mixer and gate configurations for the long-term benchmarks.}
\label{tab:maro-cfg-gates}
\resizebox{\columnwidth}{!}{%
\begin{tabular}{llrrrrrr}
\toprule
Dataset & $H$ & $\kappa$ & $\kappa/(C\!-\!1)$ & $d_g$ & $\beta$ & $g_{\mathrm{ch}}$ init & $g_r$ init \\
\midrule
ETTh1 & 96 & 2 & 0.333 & 16 & 1.00 & $8.93\times 10^{-4}$ & 0.200 \\
ETTh1 & 192 & 6 & 1.000 & 8 & 0.00 & $1.38\times 10^{-2}$ & 0.010 \\
ETTh1 & 336 & 6 & 1.000 & 32 & 1.00 & $4.72\times 10^{-2}$ & 0.100 \\
ETTh1 & 720 & 2 & 0.333 & 16 & 1.00 & $3.66\times 10^{-4}$ & 0.010 \\
\midrule
ETTh2 & 96 & 5 & 0.833 & 128 & 2.00 & $3.99\times 10^{-3}$ & 0.050 \\
ETTh2 & 192 & 5 & 0.833 & 4 & 0.50 & $6.16\times 10^{-3}$ & 0.010 \\
ETTh2 & 336 & 6 & 1.000 & 64 & 1.00 & $2.52\times 10^{-3}$ & 0.100 \\
ETTh2 & 720 & 5 & 0.833 & 8 & 2.00 & $1.16\times 10^{-3}$ & 0.010 \\
\midrule
ETTm1 & 96 & 5 & 0.833 & 128 & 1.00 & $2.30\times 10^{-2}$ & 0.010 \\
ETTm1 & 192 & 5 & 0.833 & 64 & 1.00 & $2.24\times 10^{-3}$ & 0.050 \\
ETTm1 & 336 & 5 & 0.833 & 32 & 1.00 & $5.64\times 10^{-4}$ & 0.100 \\
ETTm1 & 720 & 2 & 0.333 & 4 & 2.00 & $2.70\times 10^{-3}$ & 0.200 \\
\midrule
ETTm2 & 96 & 5 & 0.833 & 4 & 0.00 & $7.37\times 10^{-3}$ & 0.050 \\
ETTm2 & 192 & 2 & 0.333 & 64 & 2.00 & $8.65\times 10^{-4}$ & 0.010 \\
ETTm2 & 336 & 5 & 0.833 & 8 & 1.00 & $2.23\times 10^{-4}$ & 0.025 \\
ETTm2 & 720 & 2 & 0.333 & 16 & 2.00 & $5.48\times 10^{-3}$ & 0.200 \\
\midrule
Weather & 96 & 5 & 0.250 & 8 & 0.00 & $1.22\times 10^{-2}$ & 0.100 \\
Weather & 192 & 16 & 0.800 & 8 & 2.00 & $5.72\times 10^{-4}$ & 0.050 \\
Weather & 336 & 6 & 0.300 & 64 & 1.00 & $2.29\times 10^{-3}$ & 0.050 \\
Weather & 720 & 5 & 0.250 & 4 & 0.50 & $5.30\times 10^{-4}$ & 0.050 \\
\midrule
Electricity & 96 & 64 & 0.200 & 128 & 0.00 & $1.26\times 10^{-2}$ & 0.100 \\
Electricity & 192 & 2 & 0.006 & 4 & 2.00 & $9.37\times 10^{-4}$ & 0.025 \\
Electricity & 336 & 100 & 0.312 & 32 & 1.00 & $3.08\times 10^{-3}$ & 0.100 \\
Electricity & 720 & 100 & 0.312 & 32 & 1.00 & $4.26\times 10^{-4}$ & 0.010 \\
\midrule
Traffic & 96 & 64 & 0.074 & 32 & 0.00 & $1.06\times 10^{-4}$ & 0.025 \\
Traffic & 192 & 5 & 0.006 & 16 & 1.00 & $4.41\times 10^{-3}$ & 0.100 \\
Traffic & 336 & 32 & 0.037 & 64 & 0.50 & $1.87\times 10^{-2}$ & 0.050 \\
Traffic & 720 & 6 & 0.007 & 8 & 2.00 & $8.99\times 10^{-4}$ & 0.200 \\
\midrule
Solar & 96 & 16 & 0.118 & 128 & 1.00 & $1.60\times 10^{-4}$ & 0.050 \\
Solar & 192 & 6 & 0.044 & 4 & 0.50 & $3.85\times 10^{-4}$ & 0.050 \\
Solar & 336 & 32 & 0.235 & 128 & 2.00 & $4.61\times 10^{-2}$ & 0.100 \\
Solar & 720 & 6 & 0.044 & 16 & 0.00 & $1.85\times 10^{-4}$ & 0.010 \\
\bottomrule
\end{tabular}%
}
\end{table}

%% file: tables/table1_full.tex
\begin{table}[H]
\centering
\caption{Full long-term forecasting results. The remaining baselines are shown
in Table~\ref{tab:full-results-cont}.}
\label{tab:full-results}
\label{tab:maro-seeds}
\renewcommand{\arraystretch}{0.94}
\resizebox{\textwidth}{!}{%
\begin{tabular}{llcccccccc}
\toprule
 &  & \multicolumn{2}{c}{MARO (Ours)} & \multicolumn{2}{c}{SRSNet} & \multicolumn{2}{c}{TimeKAN} & \multicolumn{2}{c}{Amplifier} \\
\cmidrule(lr){3-4} \cmidrule(lr){5-6} \cmidrule(lr){7-8} \cmidrule(lr){9-10}
Dataset & Horizon & MSE & MAE & MSE & MAE & MSE & MAE & MSE & MAE \\
\midrule
ETTh1 & 96 & \textbf{0.354}\,{\scriptsize$\pm0.0005$} & \textbf{0.390}\,{\scriptsize$\pm0.0004$} & \underline{0.366} & \underline{0.394} & 0.370 & 0.396 & 0.373 & 0.399  \\
 & 192 & \textbf{0.391}\,{\scriptsize$\pm0.0008$} & \textbf{0.411}\,{\scriptsize$\pm0.0010$} & \underline{0.400} & \underline{0.415} & 0.403 & 0.417 & 0.414 & 0.420  \\
 & 336 & \textbf{0.398}\,{\scriptsize$\pm0.0019$} & \textbf{0.429}\,{\scriptsize$\pm0.0027$} & 0.424 & \underline{0.430} & \underline{0.420} & 0.432 & 0.442 & 0.446  \\
 & 720 & \underline{0.440}\,{\scriptsize$\pm0.0030$} & \underline{0.460}\,{\scriptsize$\pm0.0024$} & \textbf{0.426} & \textbf{0.455} & 0.442 & 0.465 & 0.455 & 0.467  \\
\cmidrule(lr){1-10}
\textit{Avg} & & \textbf{0.396} & \textbf{0.423} & \underline{0.404} & \textbf{0.423} & 0.408 & \underline{0.427} & 0.421 & 0.433  \\
\midrule
ETTh2 & 96 & \textbf{0.254}\,{\scriptsize$\pm0.0018$} & \textbf{0.324}\,{\scriptsize$\pm0.0011$} & 0.271 & 0.338 & 0.280 & 0.343 & 0.287 & 0.349  \\
 & 192 & \textbf{0.312}\,{\scriptsize$\pm0.0043$} & \textbf{0.368}\,{\scriptsize$\pm0.0028$} & \underline{0.324} & \underline{0.372} & 0.329 & 0.382 & 0.348 & 0.393  \\
 & 336 & \textbf{0.314}\,{\scriptsize$\pm0.0004$} & \textbf{0.376}\,{\scriptsize$\pm0.0002$} & \underline{0.349} & \underline{0.394} & 0.370 & 0.412 & 0.383 & 0.423  \\
 & 720 & \textbf{0.383}\,{\scriptsize$\pm0.0038$} & \textbf{0.425}\,{\scriptsize$\pm0.0032$} & \underline{0.391} & \underline{0.434} & 0.420 & 0.450 & 0.407 & 0.444  \\
\cmidrule(lr){1-10}
\textit{Avg} & & \textbf{0.316} & \textbf{0.373} & \underline{0.333} & \underline{0.384} & 0.349 & 0.396 & 0.356 & 0.402  \\
\midrule
ETTm1 & 96 & \textbf{0.287}\,{\scriptsize$\pm0.0033$} & 0.346\,{\scriptsize$\pm0.0019$} & \underline{0.288} & \textbf{0.341} & 0.290 & 0.348 & 0.292 & 0.346  \\
 & 192 & 0.330\,{\scriptsize$\pm0.0007$} & 0.368\,{\scriptsize$\pm0.0006$} & \textbf{0.326} & \textbf{0.364} & 0.332 & 0.368 & \underline{0.327} & \underline{0.365}  \\
 & 336 & 0.360\,{\scriptsize$\pm0.0038$} & \underline{0.385}\,{\scriptsize$\pm0.0012$} & 0.365 & 0.386 & \textbf{0.354} & 0.386 & 0.365 & 0.386  \\
 & 720 & 0.421\,{\scriptsize$\pm0.0009$} & \textbf{0.416}\,{\scriptsize$\pm0.0006$} & 0.426 & 0.420 & \textbf{0.401} & \underline{0.417} & 0.427 & 0.419  \\
\cmidrule(lr){1-10}
\textit{Avg} & & 0.349 & 0.379 & 0.351 & \textbf{0.377} & \textbf{0.344} & 0.379 & 0.352 & 0.379  \\
\midrule
ETTm2 & 96 & \textbf{0.163}\,{\scriptsize$\pm0.0004$} & \textbf{0.251}\,{\scriptsize$\pm0.0004$} & \underline{0.164} & \underline{0.254} & \underline{0.164} & \underline{0.254} & \underline{0.164} & \underline{0.254}  \\
 & 192 & \textbf{0.218}\,{\scriptsize$\pm0.0007$} & \textbf{0.291}\,{\scriptsize$\pm0.0002$} & \underline{0.220} & 0.296 & 0.238 & 0.300 & 0.226 & 0.300  \\
 & 336 & 0.275\,{\scriptsize$\pm0.0019$} & \underline{0.328}\,{\scriptsize$\pm0.0024$} & \textbf{0.271} & \textbf{0.327} & 0.278 & 0.331 & 0.276 & 0.331  \\
 & 720 & \underline{0.355}\,{\scriptsize$\pm0.0020$} & \underline{0.381}\,{\scriptsize$\pm0.0011$} & \textbf{0.353} & \textbf{0.380} & 0.359 & 0.387 & 0.358 & 0.388  \\
\cmidrule(lr){1-10}
\textit{Avg} & & \underline{0.253} & \textbf{0.313} & \textbf{0.252} & \underline{0.314} & 0.259 & 0.318 & 0.256 & 0.318  \\
\midrule
Weather & 96 & \textbf{0.145}\,{\scriptsize$\pm0.0008$} & \textbf{0.194}\,{\scriptsize$\pm0.0004$} & \underline{0.147} & \underline{0.198} & 0.151 & 0.205 & \underline{0.147} & 0.199  \\
 & 192 & \underline{0.191}\,{\scriptsize$\pm0.0007$} & \underline{0.240}\,{\scriptsize$\pm0.0009$} & \textbf{0.190} & 0.242 & 0.195 & 0.244 & 0.194 & 0.245  \\
 & 336 & \underline{0.242}\,{\scriptsize$\pm0.0009$} & \textbf{0.278}\,{\scriptsize$\pm0.0008$} & \textbf{0.241} & 0.282 & \underline{0.242} & 0.287 & 0.243 & 0.282  \\
 & 720 & 0.315\,{\scriptsize$\pm0.0013$} & 0.331\,{\scriptsize$\pm0.0006$} & 0.325 & 0.340 & 0.317 & 0.340 & \textbf{0.310} & \textbf{0.329}  \\
\cmidrule(lr){1-10}
\textit{Avg} & & \textbf{0.223} & \textbf{0.261} & \underline{0.225} & 0.265 & 0.226 & 0.269 & \textbf{0.223} & 0.263  \\
\midrule
Electricity & 96 & \textbf{0.126}\,{\scriptsize$\pm0.0004$} & \textbf{0.223}\,{\scriptsize$\pm0.0004$} & \underline{0.131} & \underline{0.226} & 0.135 & 0.231 & 0.132 & 0.227  \\
 & 192 & \textbf{0.145}\,{\scriptsize$\pm0.0006$} & \textbf{0.241}\,{\scriptsize$\pm0.0003$} & \underline{0.147} & \textbf{0.241} & 0.149 & \underline{0.243} & 0.149 & \underline{0.243}  \\
 & 336 & \textbf{0.162}\,{\scriptsize$\pm0.0027$} & 0.262\,{\scriptsize$\pm0.0022$} & \underline{0.163} & \textbf{0.258} & 0.165 & \underline{0.260} & 0.167 & 0.261  \\
 & 720 & \textbf{0.186}\,{\scriptsize$\pm0.0054$} & \textbf{0.282}\,{\scriptsize$\pm0.0033$} & 0.201 & 0.291 & 0.206 & 0.297 & 0.203 & 0.292  \\
\cmidrule(lr){1-10}
\textit{Avg} & & \textbf{0.155} & \textbf{0.252} & \underline{0.160} & \underline{0.254} & 0.163 & 0.257 & 0.162 & 0.255  \\
\midrule
Traffic & 96 & \textbf{0.340}\,{\scriptsize$\pm0.0031$} & \textbf{0.252}\,{\scriptsize$\pm0.0013$} & \underline{0.361} & \underline{0.254} & 0.388 & 0.269 & 0.391 & 0.277  \\
 & 192 & \textbf{0.351}\,{\scriptsize$\pm0.0004$} & \textbf{0.254}\,{\scriptsize$\pm0.0006$} & \underline{0.380} & \underline{0.263} & 0.411 & 0.286 & 0.405 & 0.283  \\
 & 336 & \textbf{0.380}\,{\scriptsize$\pm0.0022$} & \textbf{0.269}\,{\scriptsize$\pm0.0027$} & \underline{0.392} & \underline{0.270} & 0.425 & 0.284 & 0.416 & 0.290  \\
 & 720 & \textbf{0.418}\,{\scriptsize$\pm0.0026$} & \textbf{0.285}\,{\scriptsize$\pm0.0018$} & \underline{0.434} & \underline{0.293} & 0.455 & 0.302 & 0.436 & 0.312  \\
\cmidrule(lr){1-10}
\textit{Avg} & & \textbf{0.372} & \textbf{0.265} & \underline{0.391} & \underline{0.270} & 0.419 & 0.285 & 0.412 & 0.290  \\
\midrule
Solar & 96 & \underline{0.169}\,{\scriptsize$\pm0.0107$} & \underline{0.223}\,{\scriptsize$\pm0.0137$} & \textbf{0.167} & \textbf{0.222} & 0.187 & 0.255 & 0.175 & 0.237  \\
 & 192 & \underline{0.190}\,{\scriptsize$\pm0.0056$} & \underline{0.251}\,{\scriptsize$\pm0.0080$} & \textbf{0.182} & \textbf{0.237} & 0.194 & 0.265 & 0.198 & 0.259  \\
 & 336 & \textbf{0.187}\,{\scriptsize$\pm0.0035$} & \textbf{0.245}\,{\scriptsize$\pm0.0013$} & \underline{0.188} & \textbf{0.245} & 0.203 & 0.264 & 0.213 & \underline{0.259}  \\
 & 720 & \underline{0.198}\,{\scriptsize$\pm0.0019$} & 0.261\,{\scriptsize$\pm0.0061$} & \textbf{0.195} & \textbf{0.251} & 0.209 & 0.269 & 0.222 & 0.369  \\
\cmidrule(lr){1-10}
\textit{Avg} & & \underline{0.186} & \underline{0.245} & \textbf{0.183} & \textbf{0.238} & 0.198 & 0.263 & 0.202 & 0.281  \\
\midrule
\multicolumn{2}{l}{\# Best / Total}  & \textbf{20/32} & \textbf{20/32} & \underline{9/32} & \underline{11/32} & 2/32 & 0/32 & 1/32 & 1/32 \\
\bottomrule
\end{tabular}%
}
\end{table}

\begin{table}[H]
\centering
\caption{Full long-term forecasting results continued from
Table~\ref{tab:full-results}.}
\label{tab:full-results-cont}
\renewcommand{\arraystretch}{0.92}
\resizebox{\textwidth}{!}{%
\begin{tabular}{llcccccccc}
\toprule
 &  & \multicolumn{2}{c}{MARO (Ours)} & \multicolumn{2}{c}{iTransformer} & \multicolumn{2}{c}{TimeMixer} & \multicolumn{2}{c}{PatchTST} \\
\cmidrule(lr){3-4} \cmidrule(lr){5-6} \cmidrule(lr){7-8} \cmidrule(lr){9-10}
Dataset & Horizon & MSE & MAE & MSE & MAE & MSE & MAE & MSE & MAE \\
\midrule
ETTh1 & 96 & \textbf{0.354}\,{\scriptsize$\pm0.0005$} & \textbf{0.390}\,{\scriptsize$\pm0.0004$} & 0.386 & 0.405 & 0.372 & 0.401 & 0.377 & 0.397  \\
 & 192 & \textbf{0.391}\,{\scriptsize$\pm0.0008$} & \textbf{0.411}\,{\scriptsize$\pm0.0010$} & 0.430 & 0.435 & 0.413 & 0.430 & 0.409 & 0.425  \\
 & 336 & \textbf{0.398}\,{\scriptsize$\pm0.0019$} & \textbf{0.429}\,{\scriptsize$\pm0.0027$} & 0.450 & 0.452 & 0.438 & 0.450 & 0.431 & 0.444  \\
 & 720 & \underline{0.440}\,{\scriptsize$\pm0.0030$} & \underline{0.460}\,{\scriptsize$\pm0.0024$} & 0.495 & 0.487 & 0.483 & 0.483 & 0.457 & 0.477  \\
\cmidrule(lr){1-10}
\textit{Avg} & & \textbf{0.396} & \textbf{0.423} & 0.440 & 0.444 & 0.426 & 0.441 & 0.418 & 0.435  \\
\midrule
ETTh2 & 96 & \textbf{0.254}\,{\scriptsize$\pm0.0018$} & \textbf{0.324}\,{\scriptsize$\pm0.0011$} & 0.292 & 0.347 & \underline{0.270} & 0.342 & 0.274 & \underline{0.337}  \\
 & 192 & \textbf{0.312}\,{\scriptsize$\pm0.0043$} & \textbf{0.368}\,{\scriptsize$\pm0.0028$} & 0.348 & 0.384 & 0.349 & 0.387 & 0.348 & 0.384  \\
 & 336 & \textbf{0.314}\,{\scriptsize$\pm0.0004$} & \textbf{0.376}\,{\scriptsize$\pm0.0002$} & 0.372 & 0.407 & 0.367 & 0.410 & 0.377 & 0.416  \\
 & 720 & \textbf{0.383}\,{\scriptsize$\pm0.0038$} & \textbf{0.425}\,{\scriptsize$\pm0.0032$} & 0.424 & 0.444 & 0.401 & 0.436 & 0.406 & 0.441  \\
\cmidrule(lr){1-10}
\textit{Avg} & & \textbf{0.316} & \textbf{0.373} & 0.359 & 0.395 & 0.346 & 0.393 & 0.351 & 0.394  \\
\midrule
ETTm1 & 96 & \textbf{0.287}\,{\scriptsize$\pm0.0033$} & 0.346\,{\scriptsize$\pm0.0019$} & \textbf{0.287} & \underline{0.342} & 0.293 & 0.345 & 0.289 & 0.343  \\
 & 192 & 0.330\,{\scriptsize$\pm0.0007$} & 0.368\,{\scriptsize$\pm0.0006$} & 0.331 & 0.371 & 0.335 & 0.372 & 0.329 & 0.368  \\
 & 336 & 0.360\,{\scriptsize$\pm0.0038$} & \underline{0.385}\,{\scriptsize$\pm0.0012$} & \underline{0.358} & \textbf{0.384} & 0.368 & 0.386 & 0.362 & 0.390  \\
 & 720 & 0.421\,{\scriptsize$\pm0.0009$} & \textbf{0.416}\,{\scriptsize$\pm0.0006$} & \underline{0.412} & \textbf{0.416} & 0.426 & \underline{0.417} & 0.416 & 0.423  \\
\cmidrule(lr){1-10}
\textit{Avg} & & 0.349 & 0.379 & \underline{0.347} & \underline{0.378} & 0.355 & 0.380 & 0.349 & 0.381  \\
\midrule
ETTm2 & 96 & \textbf{0.163}\,{\scriptsize$\pm0.0004$} & \textbf{0.251}\,{\scriptsize$\pm0.0004$} & 0.168 & 0.262 & 0.165 & 0.256 & 0.165 & 0.255  \\
 & 192 & \textbf{0.218}\,{\scriptsize$\pm0.0007$} & \textbf{0.291}\,{\scriptsize$\pm0.0002$} & 0.224 & 0.295 & 0.225 & 0.298 & 0.221 & \underline{0.293}  \\
 & 336 & 0.275\,{\scriptsize$\pm0.0019$} & \underline{0.328}\,{\scriptsize$\pm0.0024$} & \underline{0.274} & 0.330 & 0.277 & 0.332 & 0.276 & \textbf{0.327}  \\
 & 720 & \underline{0.355}\,{\scriptsize$\pm0.0020$} & \underline{0.381}\,{\scriptsize$\pm0.0011$} & 0.367 & 0.385 & 0.360 & 0.387 & 0.362 & \underline{0.381}  \\
\cmidrule(lr){1-10}
\textit{Avg} & & \underline{0.253} & \textbf{0.313} & 0.258 & 0.318 & 0.256 & 0.318 & 0.256 & \underline{0.314}  \\
\midrule
Weather & 96 & \textbf{0.145}\,{\scriptsize$\pm0.0008$} & \textbf{0.194}\,{\scriptsize$\pm0.0004$} & 0.150 & 0.207 & 0.151 & \underline{0.198} & 0.150 & 0.200  \\
 & 192 & \underline{0.191}\,{\scriptsize$\pm0.0007$} & \underline{0.240}\,{\scriptsize$\pm0.0009$} & 0.200 & 0.248 & \underline{0.191} & 0.242 & \underline{0.191} & \textbf{0.239}  \\
 & 336 & \underline{0.242}\,{\scriptsize$\pm0.0009$} & \textbf{0.278}\,{\scriptsize$\pm0.0008$} & 0.252 & 0.287 & 0.244 & 0.280 & \underline{0.242} & \underline{0.279}  \\
 & 720 & 0.315\,{\scriptsize$\pm0.0013$} & 0.331\,{\scriptsize$\pm0.0006$} & 0.320 & 0.336 & 0.316 & 0.331 & \underline{0.312} & \underline{0.330}  \\
\cmidrule(lr){1-10}
\textit{Avg} & & \textbf{0.223} & \textbf{0.261} & 0.230 & 0.269 & \underline{0.225} & \underline{0.262} & \textbf{0.223} & \underline{0.262}  \\
\midrule
Electricity & 96 & \textbf{0.126}\,{\scriptsize$\pm0.0004$} & \textbf{0.223}\,{\scriptsize$\pm0.0004$} & 0.134 & 0.230 & 0.153 & 0.256 & 0.143 & 0.247  \\
 & 192 & \textbf{0.145}\,{\scriptsize$\pm0.0006$} & \textbf{0.241}\,{\scriptsize$\pm0.0003$} & 0.154 & 0.250 & 0.168 & 0.269 & 0.158 & 0.260  \\
 & 336 & \textbf{0.162}\,{\scriptsize$\pm0.0027$} & 0.262\,{\scriptsize$\pm0.0022$} & 0.169 & 0.265 & 0.189 & 0.281 & 0.168 & 0.267  \\
 & 720 & \textbf{0.186}\,{\scriptsize$\pm0.0054$} & \textbf{0.282}\,{\scriptsize$\pm0.0033$} & \underline{0.194} & \underline{0.288} & 0.228 & 0.320 & 0.214 & 0.307  \\
\cmidrule(lr){1-10}
\textit{Avg} & & \textbf{0.155} & \textbf{0.252} & 0.162 & 0.258 & 0.184 & 0.281 & 0.170 & 0.270  \\
\midrule
Traffic & 96 & \textbf{0.340}\,{\scriptsize$\pm0.0031$} & \textbf{0.252}\,{\scriptsize$\pm0.0013$} & 0.363 & 0.265 & 0.369 & 0.256 & 0.370 & 0.262  \\
 & 192 & \textbf{0.351}\,{\scriptsize$\pm0.0004$} & \textbf{0.254}\,{\scriptsize$\pm0.0006$} & 0.384 & 0.273 & 0.400 & 0.271 & 0.386 & 0.269  \\
 & 336 & \textbf{0.380}\,{\scriptsize$\pm0.0022$} & \textbf{0.269}\,{\scriptsize$\pm0.0027$} & 0.396 & 0.277 & 0.407 & 0.272 & 0.396 & 0.275  \\
 & 720 & \textbf{0.418}\,{\scriptsize$\pm0.0026$} & \textbf{0.285}\,{\scriptsize$\pm0.0018$} & 0.445 & 0.308 & 0.462 & 0.316 & 0.435 & 0.295  \\
\cmidrule(lr){1-10}
\textit{Avg} & & \textbf{0.372} & \textbf{0.265} & 0.397 & 0.280 & 0.409 & 0.278 & 0.396 & 0.275  \\
\midrule
Solar & 96 & \underline{0.169}\,{\scriptsize$\pm0.0107$} & \underline{0.223}\,{\scriptsize$\pm0.0137$} & 0.174 & 0.229 & 0.180 & 0.233 & 0.170 & 0.234  \\
 & 192 & \underline{0.190}\,{\scriptsize$\pm0.0056$} & \underline{0.251}\,{\scriptsize$\pm0.0080$} & 0.205 & 0.270 & 0.201 & 0.259 & 0.204 & 0.302  \\
 & 336 & \textbf{0.187}\,{\scriptsize$\pm0.0035$} & \textbf{0.245}\,{\scriptsize$\pm0.0013$} & 0.216 & 0.282 & 0.214 & 0.272 & 0.212 & 0.293  \\
 & 720 & \underline{0.198}\,{\scriptsize$\pm0.0019$} & 0.261\,{\scriptsize$\pm0.0061$} & 0.211 & \underline{0.260} & 0.218 & 0.278 & 0.215 & 0.307  \\
\cmidrule(lr){1-10}
\textit{Avg} & & \underline{0.186} & \underline{0.245} & 0.201 & 0.260 & 0.203 & 0.260 & 0.200 & 0.284  \\
\midrule
\multicolumn{2}{l}{\# Best / Total}  & \textbf{20/32} & \textbf{20/32} & 1/32 & 2/32 & 0/32 & 0/32 & 0/32 & 2/32 \\
\bottomrule
\end{tabular}%
}
\end{table}

%% file: tables/table_graph_ablation.tex
\begin{table}[H]
\centering
\caption{Channel-topology controls. Entries are test MSE over three seeds and
bold marks the lowest mean in each row.}
\label{tab:maro-graph-ablation}
\resizebox{\columnwidth}{!}{%
\begin{tabular}{lccc}
\toprule
Cell & Learned & Random fixed & Correlation fixed \\
\midrule
Electricity $H{=}96$  & $.12724\pm.00025$ & $.12679\pm.00005$ & $\mathbf{.12579\pm.00073}$ \\
Electricity $H{=}336$ & $\mathbf{.16202\pm.00162}$ & $.16514\pm.00285$ & $.16738\pm.00266$ \\
Solar $H{=}96$        & $.16597\pm.00261$ & $.16466\pm.00560$ & $\mathbf{.16386\pm.00200}$ \\
Solar $H{=}336$       & $\mathbf{.18707\pm.00277}$ & $.19922\pm.00190$ & $.18878\pm.00098$ \\
Traffic $H{=}96$      & $.34486\pm.00584$ & $\mathbf{.33564\pm.00056}$ & $.33966\pm.00018$ \\
Traffic $H{=}336$     & $.38336\pm.00403$ & $\mathbf{.38083\pm.00140}$ & $.38601\pm.00030$ \\
\bottomrule
\end{tabular}%
}
\end{table}

%% file: tables/table_gate_screen.tex
\begin{table}[H]
\centering
\caption{Gate controls. Entries are three-seed mean MSE changes from the
corresponding baseline and negative is better.}
\label{tab:maro-gate-screen}
\resizebox{\columnwidth}{!}{%
\begin{tabular}{lrrrr}
\toprule
Cell & Frozen $\Delta$ & LayerScale $\Delta$ & Wide $\Delta$ & Separate ctx.\ gate $\Delta$ \\
\midrule
ETTh1 $H{=}96$ & $+.000010$ & $-.000031$ & $-.000002$ & $+.000015$ \\
ETTh1 $H{=}336$ & $+.000009$ & $+.000040$ & $-.000015$ & $+.000052$ \\
ETTh2 $H{=}96$ & $-.000388$ & $+.000350$ & $+.000494$ & $+.000546$ \\
ETTh2 $H{=}336$ & $-.000015$ & $+.000055$ & $-.000011$ & $-.000006$ \\
Weather $H{=}96$ & $+.000863$ & $+.000936$ & $+.001294$ & $+.000671$ \\
Weather $H{=}336$ & $-.000307$ & $-.000263$ & $-.000241$ & $-.000052$ \\
\bottomrule
\end{tabular}%
}
\end{table}